\documentclass[journal]{IEEEtran}
\usepackage{graphicx}
\usepackage{hyperref}
\usepackage{multirow}
\usepackage{amsmath}
\usepackage{subcaption}
\usepackage{amsmath}
\usepackage{amssymb}
\usepackage{booktabs}
\usepackage{tikz}
\usepackage{footnote}
\usetikzlibrary{positioning,shapes}
\usepackage{xcolor}
\usepackage{pifont}
\newcommand{\cmark}{\ding{51}}%
\newcommand{\xmark}{\ding{55}}%
\usepackage{cite}
\usepackage{colortbl}
\definecolor{gray}{gray}{0.9}

\begin{document}
\title{DDPM-CD: Denoising Diffusion Probabilistic Models as Feature Extractors for Change Detection}

\author{Wele Gedara Chaminda Bandara,~\IEEEmembership{Student Member,~IEEE,} Nithin Gopalakrishnan Nair,~\IEEEmembership{Student Member,~IEEE,} and Vishal M. Patel,~\IEEEmembership{Senior Member,~IEEE}
\thanks{The authors are with the Department
of Electrical and Computer Engineering, Johns Hopkins University, Baltimore,
MD, 21218 USA.\\ e-mail: \texttt{\{wbandar1, ngopala2, vpatel36\}@jhu.edu}}\\
Project page: \href{https://github.com/wgcban/ddpm-cd}{\color{blue}https://github.com/wgcban/ddpm-cd}
}

\markboth{IEEE Transactions on Geoscience and Remote Sensing}%
{Shell \MakeLowercase{\textit{et al.}}: Bare Demo of IEEEtran.cls for Journals}
\maketitle

\begin{abstract}
Remote sensing change detection is crucial for understanding the dynamics of our planet's surface, facilitating the monitoring of environmental changes, evaluating human impact, predicting future trends, and supporting decision-making. In this work, we introduce a novel approach for change detection that can leverage off-the-shelf, unlabeled remote sensing images in the training process by pre-training a Denoising Diffusion Probabilistic Model (DDPM) – a class of generative models used in image synthesis. DDPMs learn the training data distribution by gradually converting training images into a Gaussian distribution using a Markov chain. During inference (i.e., sampling), they can generate a diverse set of samples closer to the training distribution, starting from Gaussian noise, achieving state-of-the-art image synthesis results. However, in this work, our focus is not on image synthesis but on utilizing it as a pre-trained feature extractor for the downstream application of change detection. Specifically, we fine-tune a lightweight change classifier utilizing the feature representations produced by the pre-trained DDPM alongside change labels. Experiments conducted on the LEVIR-CD, WHU-CD, DSIFN-CD, and CDD datasets demonstrate that the proposed DDPM-CD method significantly outperforms the existing state-of-the-art change detection methods in terms of F1 score, IoU, and overall accuracy, highlighting the pivotal role of pre-trained DDPM as a feature extractor for downstream applications. We have made both the code and pre-trained models available at \href{https://github.com/wgcban/ddpm-cd}{\color{blue}https://github.com/wgcban/ddpm-cd}
\end{abstract}


\IEEEpeerreviewmaketitle

\begin{figure*}[h!]
    \centering
    \includegraphics[width=\linewidth]{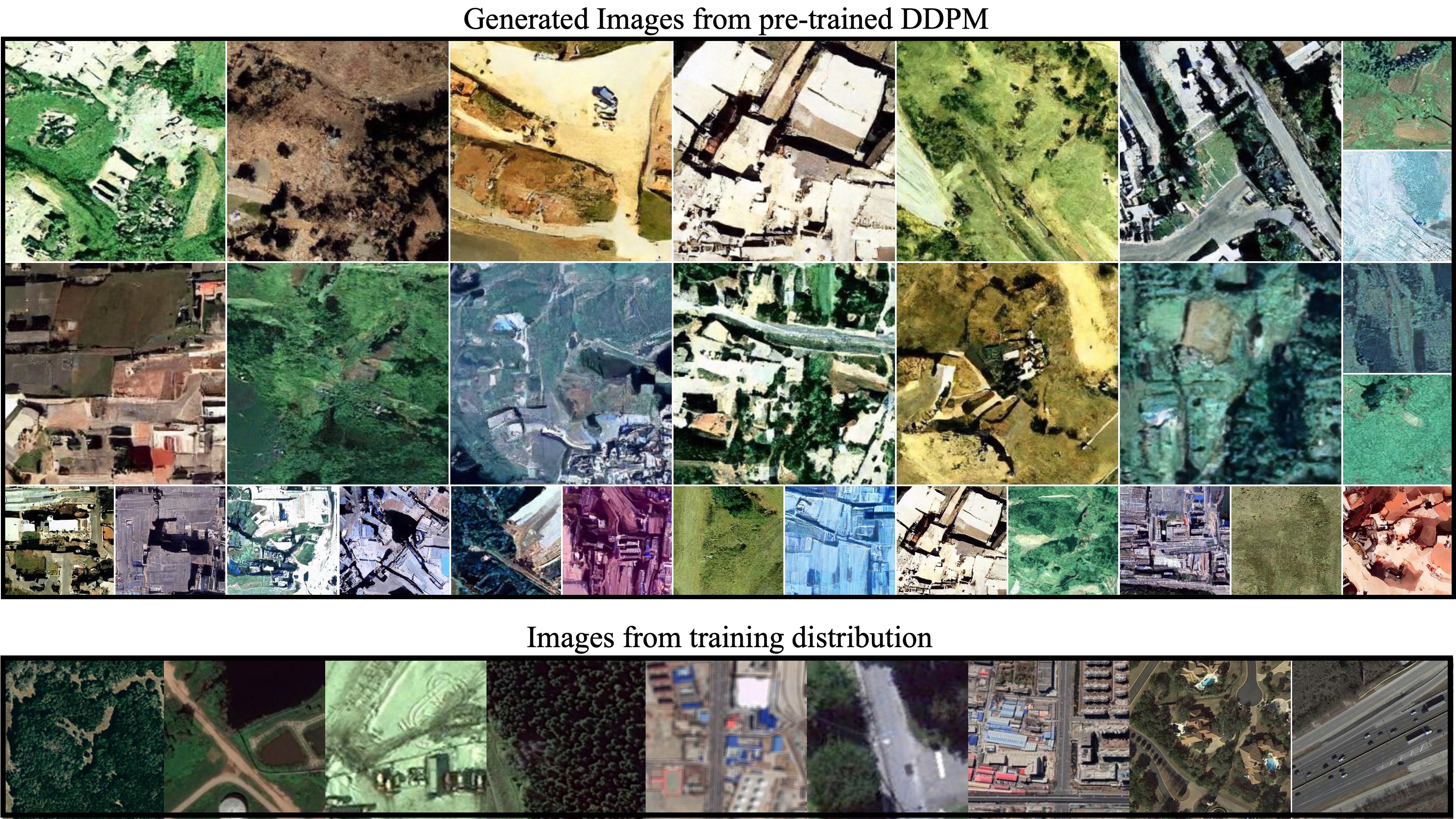}
        \caption{\textbf{Images sampled from the DDPM model pre-trained on off-the-shelf remote sensing images.} The generated images exhibit common objects typically observed in real remote sensing imagery, including buildings, trees, roads, vegetation, water surfaces, etc. This showcases the remarkable capability of diffusion models to grasp essential semantics from the training dataset. Although our primary focus isn't image synthesis, we explore the effectiveness of DDPM as a feature extractor for change detection.}
    \label{fig:samples}
\end{figure*}

\section{Introduction}
Remote sensing change detection involves identifying changes on the Earth's surface by analyzing images captured at different points in time~\cite{khelifi2020deep, bai2023deep}. Change detection is useful across multiple applications in remote sensing, serving purposes such as monitoring climate shifts~\cite{liu2016review}, assessing natural disasters~\cite{hamidi2023fast,haque2023change,he2023cross,casagli2023landslide}, urban planning~\cite{ridd1998comparison, bolorinos2020consumption}, agricultural surveys~\cite{kaur2023framework}, land cover observations~\cite{coops2023framework, thien2023detection, seyam2023identifying}, and aiding in military reconnaissance~\cite{tueller1998emerging, juntakut_Aemlaor_Jantakat_2022}. However, detecting changes in remote sensing imagery poses a complex challenge due to various factors like image resolution disparities~\cite{benediktsson2012very}, noise interference~\cite{al2010comparative,landgrebe1986noise}, registration errors~\cite{inglada2006analysis,bentoutou2005automatic}, illumination changes between images~\cite{wan2018illumination, liu2014illumination}, and the intricate nature of landscapes. Recently,   deep learning-based change detection methods have substantially improved the accuracy and efficiency of change detection.

Many deep learning-based methods proposed in the literature for remote sensing change detection heavily rely on supervised-learning models~\cite{chen2022egde, shi2020change, fang2023changer, wu2023fully, lv2023novel,aleissaee2023transformers,chen2023special}. These models necessitate extensive datasets with precisely annotated changes, a requirement that is both expensive and time-consuming to fulfill. Moreover, their dependence on labeled data restricts their ability to leverage the vast amounts of freely available, off-the-shelf, unlabeled remote sensing imagery, leading to an under-utilization of available resources. Hence, the development of change detection methods that can effectively harness both unlimited unlabeled and limited labeled remote sensing images is important and timely.

Semi-supervised~\cite{hao2023semi,bandara2022revisiting, kim2021recent} and self-supervised learning~\cite{ericsson2022self, wang2022empirical} have emerged as prominent approaches which are capable of integrating unlabeled data into remote sensing change detection. In semi-supervised learning, labeled data guides the network using supervised cross-entropy loss for change detection, while unlabeled data strengthens the robustness of the learned representations by assuming concepts like the cluster~\cite{chapelle2005semi}, continuity/smoothness~\cite{kim2021recent}, and entropy~\cite{grandvalet2004semi} assumptions. On the contrary, self-supervised learning adopts a two-stage paradigm. Firstly, the model learns to predict certain features of the input images (pretext task) ~\cite{ericsson2022self} and then leverages this pre-trained knowledge for downstream tasks like change detection. The transferability and generalizability of learned representations, flexibility of adapting the pre-trained backbone for multiple downstream applications, and strong performance improvements have made self-supervised learning more attractive choice to leverage information from unlabeled data compared to semi-supervised learning.

Inspired by self-supervised learning techniques developed for natural language and image domains~\cite{ericsson2022self}, several methods have adapted these techniques to the remote sensing domain~\cite{remote_sensing_pretraining, 9844015, jiang2023self, aghdami2023effect}. For instance, in \cite{vincenzi2021color}, representations were learned from satellite imagery by reconstructing visible colors from high-dimensionality spectral bands similar to image colorization in natural images~\cite{zhang2016colorful} to learn representations. Tile2Vec \cite{jean2019tile2vec} extended the distributional hypothesis from natural language to spatial data by learning representations at the remote sensing tile level. Recent works like GeoMoCo~\cite{ayush2021geography} and SeCo~\cite{manas2021seasonal} integrated contrastive learning and temporal/positional invariance into remote sensing pre-training to enhance representations for downstream tasks. More recently, RingMo~\cite{9844015} leverages the benefits of masked image modeling (MIM) ~\cite{Xie_2022_CVPR, NEURIPS2022_416f9cb3, Bandara_2023_CVPR} to learn representations in remote sensing images useful across a variety of downstream applications.

In contrast to the aforementioned self-supervised learning techniques, this paper explores the utilization of the Denoising Diffusion Probabilistic Models (DDPMs)~\cite{ho2020denoising, nichol2021improved, yang2023diffusion,ho2020denoising, dhariwal2021diffusion} as a feature extractor for remote sensing images to facilitate accurate change detection. DDPMs were originally proposed for image synthesis tasks in generative AI, inspired by non-equilibrium thermodynamics. They define a Markov chain of diffusion steps to gradually introduce random noise to data and then learn to reverse the diffusion process to construct the desired data samples from the noise. While DDPMs excel in generating images resembling real-world scenes~\cite{nair2023unite, zhang2023adding, rombach2022high}, in this work, our focus lies in using them as feature extractors for pre-change and post-change images and evaluating their usefulness in change detection. Our empirical results demonstrate that when a lightweight change classifier is fine-tuned on top of feature representations obtained from a pre-trained DDPM, it outperforms state-of-the-art change detection methods by a significant margin. This empirical demonstration underscores its exceptional ability to produce discriminative feature representations that are useful for change detection. Moreover, as DDPMs are trained with noisy images, they yield more robust and generalized representations. These representations are more invariant to various environmental perturbations such as seasonal changes, illumination shifts, noise/blur introduced during satellite image capture, and errors in image preprocessing like registration errors.

In summary, we make the following contributions:
\begin{enumerate}
    \item We propose a novel self-supervised representation learning approach for remote sensing images which learns robust features from diffusion process in DDPMs.
    \item We empirically demonstrate that the DDPMs can produce robust and discriminative representations from remote sensing images that can be highly useful in remote sensing change detection.
    \item We demonstrate that fine-tuning a light-weight change detection classifier on top of multi-scale, multi-timestep feature representations obtained from a pre-trained DDPM works very well for change detection.
    \item  We conduct extensive experiments on four publically available change detection datasets namely, LEVIR-CD, WHU-CD, DSIFN-CD, and CDD, and demonstrate that the proposed DDPM-CD achieves better change detection than the existing state-of-the-art change detection methods by a significant margin.
\end{enumerate}

The rest of this paper is organized as follows: Section \ref{sec:related_work} presents an overview of DDPMs as well as existing change detection methods. In Section \ref{sec:method}, we delve into the detailed description of proposed DDPM-CD. Section \ref{sec:experiments} elaborates on our experimental setup and compares the results obtained by DDPM-CD with existing approaches, both qualitatively and quantitatively. In Section \ref{sec:ablation}, we conduct an ablation experiment to discern the scenarios where the pre-trained diffusion probabilistic model yields the best representations for change detection. Finally, in Section \ref{sec:conclusion}, we present our conclusions and future work.
\color{black}

\section{Related Work}
\label{sec:related_work}
\subsection{Denosing Diffusion Probabilistic Models (DDPMs)}
\label{sec:related_work_diffusion}
Denosing diffusion probabilistic models~\cite{ho2020denoising,sohl2015deep,saharia2021image, nair2022image, perera2022sar} belong to the class of generative models. Once trained, we can use the model to generate high-quality images closer to the training data distribution. Unlike other generative models such as Generative Adversarial Networks (GANs)~\cite{goodfellow2020generative, creswell2018generative}, Variational Autoencoders (VAEs)~\cite{doersch2016tutorial}, autoregressive models~\cite{esser2021imagebart}, and flows~\cite{liu2019conditional}, diffusion models are efficient to train and have been widely adopted in several applications, including super-resolution~\cite{saharia2021image}, debluring~\cite{whang2021deblurring}, segmentation~\cite{wolleb2021diffusion, jiang2018difnet, baranchuk2021label}, image colorization~\cite{song2020score}, inpainting~\cite{song2020score}, and semantic editing~\cite{rombach2021high}. However, there is still no demonstration of the application of diffusion models in remote sensing applications despite their high potential and great success in other machine vision applications. In this work, we demonstrate that remote sensing change detectioncan can also benefit from diffusion models, opening a new direction to explore remote sensing applications in the future. The following sections provide a brief overview of the diffusion framework.

\begin{figure}[tb!]
    \centering
    \includegraphics[width=\linewidth]{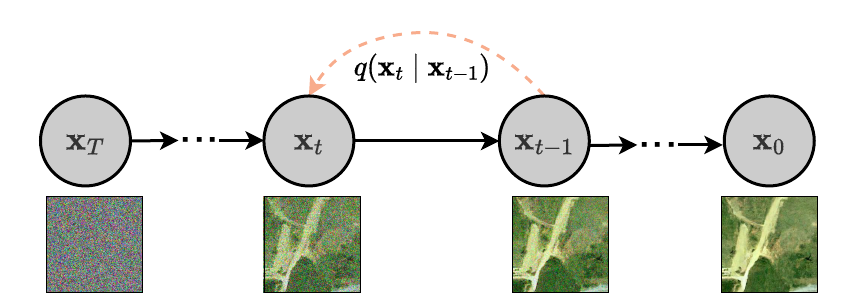}
    \caption{\textbf{Diffusion model as a directed graphical model ~\cite{ho2020denoising}.}}
    \label{fig:graph}
\end{figure}

\paragraph*{\bf Gaussian diffusion process}  Diffusion models learn the training data distribution $p(x_0)$ by performing variational inference on a Markovian process with $T$ number of timesteps. The diffusion process consists of a forward and a backward process. In the forward process, we gradually add Gaussian noise to the clean image $x_0 \sim p(x_0)$ until the image is completely destroyed, resulting in an isotropic gaussian distribution $\mathcal{N}(\boldsymbol{0}, \boldsymbol{I})$.  The noising operation at each time step $t$ in the forward process is defined as  follows:
\begin{equation}
    q(x_t|x_{t-1}) = \mathcal{N}(x_t; \sqrt{\alpha_t} x_{t-1}, (1-\alpha_t) \boldsymbol{I}),
    \label{eq:q_sample}
\end{equation}
where $(x_0, x_1, \cdots, x_T)$ denotes the $T$-step Markov chain (see Fig. \ref{fig:graph}), $\alpha_{1:T} = (\alpha_1, \cdots, \alpha_T)$ is the noise schedule that controls the variance of noise added at each step. In the reverse process, a sequence of small denoising operations is performed using a neural network to obtain back the original image. During each step of the reverse process, we perform a denoising operation to the image $x_t$ to obtain back $x_{t-1}$. For this, a neural network $f_{\theta}$ is used  to model the parameters of the reverse distribution $p(x_{t-1}|x_{t}):= \mathcal{N}(x_{t-1}:\mu_{\theta}(x_t,t), \Sigma_{\theta}(x_t,t))$. The parameters for this network $\theta$ are obtained by minimizing the KL divergence between the forward and reverse distributions over all timesteps.  During optimization, sampling from the distribution $q(x_t|x_{t-1})$ requires knowledge of $x_{t-1}$, which requires a sequence of operations to be computed as per equation \ref{eq:from_start}. The marginal distribution of $x_t$ given $x_0$ can be derived by marginalizing out the intermediate latent variables as:
\begin{equation}
    q(x_t|x_0) = \mathcal{N}(x_t; \sqrt{\gamma_t} x_0 , (1-\gamma_t) \boldsymbol{I}),
    \label{eq:from_start}
\end{equation}
where $\gamma_t = \prod_{i=1}^t \alpha_i$. Hence simplifying the calculation procedure. Furthermore, Sohl \textit{et al.}\cite{sohl2015deep} have shown minimizing the KL divergence between the forward and reverse distributions could be simplified by using the  posterior distribution $q(x_{t}|x_{t-1},x_0)$ instead of the  forward distribution $q(x_{t}|q{x_{t}]-1]})$. Under the Markovian assumptions, using equations (\ref{eq:q_sample}) and (\ref{eq:from_start}),  the posterior distribution can be derived as:
\begin{equation}
    q(x_{t-1}|x_t, x_0) = \mathcal{N}(x_{t-1}, \mu, \sigma^2 \boldsymbol{I}),
\end{equation}
where
\begin{align}
    \mu(x_t, x_0) &= \frac{\sqrt{\gamma_{t-1}} (1-\alpha_t)}{1-\gamma_t}x_0 + \frac{\sqrt{\alpha_t}(1-\gamma_{t-1})}{1-\gamma_t} x_t, \\ \sigma^2 &=  \frac{(1-\gamma_{t-1})(1-\alpha_t)}{1-\gamma_t}.
\end{align}
This posterior distribution is further utilized when parameterizing the reverse Markov chain and formulating a variational lower bound on the log-likelihood of the reverse chain. While optimizing, the covariance matrix for both the distributions $q(x_{t-1}|x_t, x_0)$ and $p(x_{t-1}|x_{t})$ is usually considered the same, and the network $f_{\theta}$ predicts the mean of the distribution.  

\paragraph*{\bf Optimizing the diffusion model} The denoising model $f_{\theta}$ takes as input the noisy image ${x_t}$:
\begin{equation}
    x_t = \sqrt{\gamma_t} x_0 + \sqrt{1-\gamma_t} \epsilon, \text{ where } \epsilon = \mathcal{N}(\boldsymbol{0}, \boldsymbol{I})
    \label{eqn:noisy_img}
\end{equation}
and the timestep $t$, and aims to recover the mean of the posterior distribution. Recent work by Ho \textit{et al}.\cite{ho2020denoising} has simplified this training objective for optimizing the parameters of the network and proposed a simplified objective defined by:
\begin{equation}
    \mathbb{E}_{x_0, \epsilon} \left\| f_{\theta}(\Tilde{x}, t ) - \epsilon \right\|_2^2, \text{ where } \epsilon = \mathcal{N}(\boldsymbol{0}, \boldsymbol{I}).
\end{equation}

\paragraph*{\bf Inference (i.e., sampling)}The inference is defined as the reverse Markovian process, which goes in the reverse direction of the forward diffusion process. The inference starts from Gaussian noise $x_T$, and $x_t$ at each timestep $t$ is iteratively  denoised to get back $x_{t-1}$ according to:
\begin{equation}
    x_{t-1} \leftarrow{} \frac{1}{\sqrt{\alpha_t}} \left( x_t - \frac{1 - \alpha_t}{\sqrt{1-\gamma_t}}f_{\theta}(x_t, t) \right) + \gamma_t \boldsymbol{z},
\end{equation}
where $\boldsymbol{z} = \mathcal{N}(\boldsymbol{0}, \boldsymbol{I})$ and  $t = T, \cdots, 1$.

\subsection{Remote Sensing Change Detection} 
\subsubsection{Classical change detection methods}
Classical change detection methods in remote sensing can be primarily categorized into three groups: (1) algebraic, (2) transformation-based, and (3) classification-based techniques.

Algebraic methods, including image differencing (ImageDiff)~\cite{ImageRatio}, image regression (ImageRegr)~\cite{ImageRegr}, image ratioing (ImageRatio)~\cite{ImageRatio}, and change vector analysis (CVA)~\cite{cva}, rely on selecting thresholds to identify altered areas. These methods, except for CVA, are relatively simple to implement but cannot provide comprehensive matrices of change information. Their reliance on threshold selection remains a significant drawback.

Transformation-based methods, such as Principal Component Analysis (PCA)\cite{dpca,PCDA}, Karhunen-Loève Transform (KT)\cite{ImageRatio}, Gramm–Schmidt (GS)\cite{ImageRatio}, Multivariate Alteration Detection (MAD)\cite{MAD}, Re-weighted Multivariate Alteration Detection (IRMAD)\cite{IRMAD}, and Chi-square transformations\cite{ImageRatio}, aim to reduce data redundancy between bands and emphasize different information in derived components. However, they often require threshold selection and encounter challenges in interpreting and labeling change information on transformed images.

Contrarily, classification-based methods like post-classification comparison~\cite{ImageRatio}, spectral–temporal combined change analysis~\cite{ImageRatio}, and expectation–maximization algorithm (EM) change detection~\cite{ImageRatio}, operate based on classified images. These methods heavily rely on the quality and quantity of training sample data to produce accurate classification results. They offer the advantage of providing change information matrices, mitigating external impacts from atmospheric and environmental differences between multi-temporal images. However, their modeling capacity and change detection quality are limited compared to modern deep learning-based approaches.

\subsubsection{Deep learning-based change detection methods}
The current research on remote sensing change detection has been significantly reshaped by deep learning owing to its powerful feature extraction ability~\cite{asokan2019change}. Initially, it was primarily based on fully convolutional neural networks (CNNs) and did not utilize any form of pre-training; instead, it solely relied on supervised learning from labeled data in an end-to-end fashion. Examples of such approaches include Fully-Convolutional Early Fusion (FC-EF)\cite{daudt2018fully}, Fully-Convolutional Siamese Concatenation (FC-Siam-conc)~\cite{daudt2018fully}, and Fully-Convolutional Siamese Difference (FC-Siam-diff)~\cite{daudt2018fully}. In the EF architecture, pre-change and post-change images are concatenated before passing them through the CNN, treating them as different color channels. In the Siamese network architecture, the encoding layers of the network are bifurcated into two streams of equal structure with shared weights, and each image is assigned to one of these streams. Subsequently, a feature difference (FC-Siam-diff) or feature concatenation (FC-Siam-conc) is applied before the final change classifier. In many cases, the Siamese difference/concatenation architecture has proven effective for change detection. Consequently, it became commonly utilized in later works for change detection purposes.

With the evolution of more potent CNN architectures such as VGG~\cite{simonyan2014very}, ResNet~\cite{he2016deep}, DenseNet~\cite{huang2017densely}, and the availability of their pre-trained models on large-scale natural image datasets like ImageNet, remote sensing methods employing transfer learning from natural images to remote sensing images have emerged. For instance, DS-IFN (deeply supervised image fusion network)~\cite{zhang2020deeply}, DASNet (dual attentive Siamese network)~\cite{9259045}, SemiCD (semi-supervised change detection)~\cite{bandara2022revisiting}, and ADS-Net (attention-based deeply supervised network)~\cite{WANG2021102348} have utilized multi-scale features from VGG16 and ResNet50 pre-trained on ImageNet to train change detection networks.

The introduction of transformer networks~\cite{vaswani2017attention}, with the core component being multi-head self-attention (MHSA)~\cite{vaswani2017attention} capable of capturing long-range context and relationships between different positions, has seen adoption in remote sensing change detection. Inspired by the Vision Transformer (ViT)~\cite{dosovitskiy2020image} approach, where the input image is divided into fixed-size patches forming tokens that are then processed by MHSA, BIT~\cite{chen2021remote} was adapted for remote sensing change detection by operating on latent feature representations obtained from ImageNet pre-trained ResNet~\cite{he2016deep}. Furthermore, a recent work, ChangeFormer~\cite{bandara2022transformer}, proposed a fully transformer network devoid of 2D convolutions for change detection, achieving superior results compared to its counterparts. Later versions of transformer networks, such as the Swin Transformer~\cite{liu2021swin}, which substitutes the global MHSA with the shiftable window MHSA (WMHSA) to significantly reduce ViT's computational overhead, have also been adopted in remote sensing change detection, as seen in SwinSUNet~\cite{zhang2022swinsunet}.

However, transformers tend to be data-hungry and typically require a well-pre-trained model to achieve better performance. Most of the previously mentioned transformer networks proposed for change detection utilize pre-trained models on natural image datasets like ImageNet~\cite{krizhevsky2012imagenet} and ADE20k~\cite{zhou2017scene}, or are randomly initialized. This is suboptimal because aerial images possess distinct characteristics creating a significant domain gap compared to natural images, including differences in view, color, texture, layout, objects, and more. To bridge this gap, these methods attempt to narrow it by further fine-tuning the pre-trained model on the remote sensing image dataset. Nevertheless, the systematic bias introduced by ImageNet pre-training has a noticeable impact on performance~\cite{wang2022empirical}.

With the emergence of large-scale aerial scene classification datasets (such as MillionAID~\cite{long2021creating}, fMoW~\cite{christie2018functional}, and BigEarthNet~\cite{sumbul2019bigearthnet}), and access to publicly available large-scale unlabeled remote sensing datasets from various Earth observation programs, it is now possible to pre-train CNN and transformer backbones on remote sensing images. However, there have been few explorations in remote sensing pre-training, and it is still not as renowned as pre-training in the natural image domain. In Geographical Knowledge-driven Representation learning (GeoKR)~\cite{li2021geographical}, global land cover products are considered as labels and a mean-teacher framework is used to alleviate the influences of imaging time and resolution differences between RS images and geographical ones. The scarcity of large-scale remote sensing datasets is mainly in terms of category labels rather than images. Hence, it is promising to develop self-supervised pre-training methods, and some related methods have been developed.

For instance, SeCo~\cite{manas2021seasonal} leverages seasonal changes to enforce consistency between positive samples, which are unique characteristics of aerial scenes. Meanwhile, in Geography-Aware Self-Supervised Learning~\cite{ayush2021geography}, temporal information and geographical location are simultaneously fused into the MoCo-V2~\cite{he2019moco,chen2020mocov2}. Moreover, exploration into remote sensing image colorization from multi-spectral images~\cite{vincenzi2021color} and spatial properties of remote sensing images~\cite{9140372} has also been conducted.

Although these self-supervised methods do not rely on labeled data during pre-training, they still use paired multi-temporal images (like SeCo~\cite{manas2021seasonal}), access to paired multi-band spectral images (as in remote sensing colorization~\cite{vincenzi2021color}), or require spatially aligned remote sensing images with known geo-locations (as in geography-aware-ssl~\cite{ayush2021geography}). This limitation restricts their ability to easily harness information from millions of off-the-shelf remote sensing images. 

Unlike existing self-supervised methods in remote sensing, our research pioneers the use of DDPM~\cite{ho2020denoising}, originally designed for image synthesis in generative AI, as a pre-training strategy for robust feature extraction from remote sensing images. This innovative pre-training approach only requires access to readily available remote sensing image datasets. Upon pre-training the DDPM, we utilize it to extract feature representations that can be leveraged to train a light-weight change detection model with annotated change images. The extraordinary capacity of DDPM to model complex training distributions more efficiently than other generative models (such as Generative Adversarial Networks (GANs), Variational Autoencoders (VAEs), etc.) enables the extraction of highly informative and compressed feature representations of a give image. Our experiments on multiple change detection datasets show that these representations obtained from pre-trained DDPM are pivotal in enhancing change detection performance significantly. 

\color{black}
\begin{figure*}[htb!]
    \centering
    \includegraphics[trim={1.2cm 0 0.5cm 0},clip, width=\linewidth]{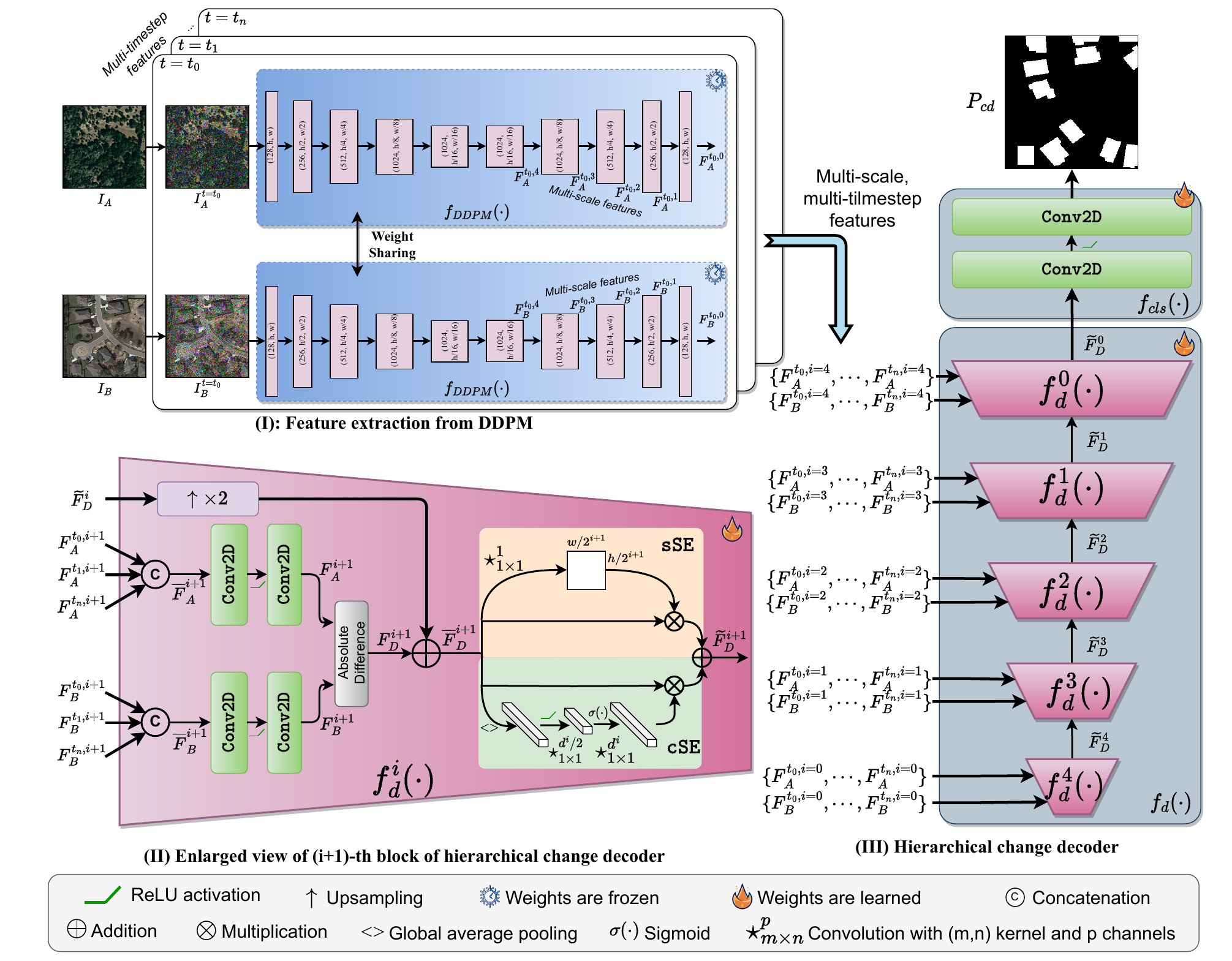}
    \caption{\textbf{Block diagrams illustrating the proposed DDPM-CD approach.} The proposed DDPM-CD involves three main steps: \textbf{(I)} Extraction of multi-scale $(i \in \{4,3,2,1,0\})$ and multi-timestep features $(t \in \{t_0, \cdots, t_n\})$ from pre-change $(I_A)$ and post-change $(I_B)$ images, denoted as $\{F_A^{t, i}\}$ and $\{F_B^{t, i}\}$, respectively. \textbf{(II)} Computation of difference feature representations at each hierarchical scale (let's say $i+1$) by the change decoder block $(f_d^{i+1}(\cdot))$, which takes multi-timestep features of pre-change and post-change images at the $i+1$-th scale, denoted as $\{F_A^{t,i+1}\}_{t=t_0}^{t=t_n}$ and $\{F_A^{t,i+1}\}_{t=t_0}^{t=t_n}$, along with the previous scale's difference feature representations $\widetilde{F}_D^{i}$ as inputs, and outputs the difference feature representations for the current scale, $\widetilde{F}_D^{i+1}$. \textbf{(III)} Cascading change decoder blocks across all spatial scales ($i=4$ to $i=0$) to form the hierarchical change decoder, represented as $f_d(\cdot)=f_d^4(f_d^3(\cdots f_d^0()))$. Finally, the output from the hierarchical change decoder $\widetilde{F}_D^0$  is fed into the change classifier $f_{cls}(\cdot)$ which predicts the change probability map $P_{cd}$.}
    \label{fig:method}
\end{figure*}

\begin{figure*}[htb!]
    \centering
    \includegraphics[width=\linewidth]{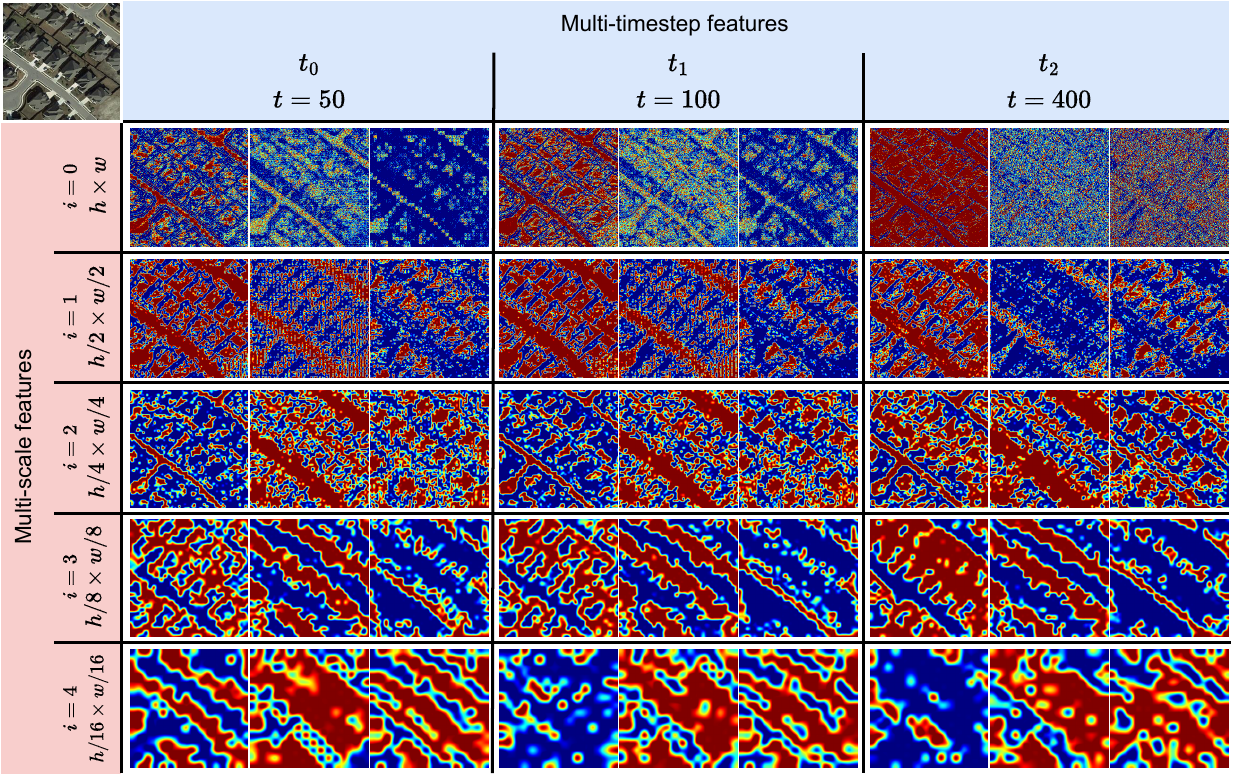}
    \caption{\textbf{Visualization of multi-scale, multi-timestep feature representations for a given input image shown in the top left corner, extracted from the pre-trained DDPM's decoder.} These multi-scale, multi-timestep feature representations are used to fine-tune a hierarchical change decoder followed by a change classifier with change labels. Here, $i$ denotes the hierarchical layer of feature maps, and $(h/2^i,w/2^i)$ denotes the (height, width) of feature representations. Additionally, $t\in[0, T]$ represents the timestep which defines the variance of noise added to the input image prior to feeding it to DDPM.}
    \label{fig:feats}
\end{figure*}

\section{Proposed Approach for Change Detection}
\label{sec:method}
The proposed DDPM-CD approach comprises of two  stages:
\begin{enumerate}
    \item \textbf{Self-supervised pre-training of DDPM} on a large collection of unlabeled remote sensing images (Sec. \ref{sec:method-pretraining}). This stage aims to learn key semantics from aerial images without relying on labeled information.
    
    \item \textbf{Utilization of pre-trained DDPM for change detection} involves fine-tuning a change detection classifier with supervised change labels (Sec. \ref{sec:method-finetuning}). This classifier leverages deep feature representations of pre-change and post-change images extracted from the decoder of the pre-trained DDPM (Sec. \ref{sec:method-featextraction}) and outputs change probability maps.
\end{enumerate}
In what follows, we describe these two steps in detail.

\subsection{Self-supervised pre-training of DDPM}
\label{sec:method-pretraining}
In order to pre-train the DDPM, we first collect a large amount ($\sim$ 500k) of 3-channel (i.e., RGB) remote sensing images of Sentinel-2~\cite{drusch2012sentinel} patches without human supervision. More specifically, we use Google Earth Engine~\cite{gorelick2017google} to process and download imagery patches from around 200K locations around the world~\cite{manas2021seasonal}, where each patch covers an earth resolution of approximately $2.65 \times 2.65$ km. Next, we train the SOTA  unconditional pixel-space U-Net ~\cite{saharia2021image} diffusion model (see Fig. \ref{fig:method}) on the collected dataset until the model converges.  Sample generated images are shown in Fig. \ref {fig:samples}.

\subsection{Multi-scale, Multi-timestep Feature Extraction from Prse-Trained DDPM}
\label{sec:method-featextraction}
The pre-trained DDPM is employed to extract feature representations from both pre-change ($I_A$) and post-change ($I_B$) images, as shown in Figure \ref{fig:method}-I. The weights of the pre-trained DDPM are kept frozen during the feature extraction.

To extract features, we begin by creating noisy images $(I_A^t, I_B^t)$ with noise variance determined by time-step $t$, akin to the process used during the pre-training of DDPM, as shown in Equation (\ref{eqn:noisy_img}):
\begin{align}
    I_A^{t} &= \sqrt{\gamma_t} I_A  + \sqrt{1-\gamma_t} \epsilon,\\
    I_B^{t} &= \sqrt{\gamma_t} I_B  + \sqrt{1-\gamma_t} \epsilon,
\end{align}
where the noise $\epsilon = \mathcal{N}(\boldsymbol{0}, \boldsymbol{I})$, and the noise variance corresponding to the timestep $t$ is given by $1-\gamma_t = 1-\prod_{i=1}^t \alpha_i$.

Next, the noisy pre-change $I_A^{t}$ and the post-change $I_B^{t}$ images are fed into the pre-trained DDPM $f_{DDPM}(\cdot)$ to obtain multi-scale $i \in \{4,3,\cdots,0\}$ features corresponding to timestep $t$:
\begin{align}
    \left \{ F_A^{t, i=0}, \cdots, F_A^{t, i=4}\right \} &= f_{DDPM}(I_A^t),\\
    \left \{ F_B^{t, i=0}, \cdots, F_B^{t, i=4} \right \} &= f_{DDPM}(I_B^t),
\end{align}
where $\left \{ F_A^{t, i=0}, \cdots, F_A^{t, i=4}\right \}$ represent the multi-scale features of the pre-change image and $\left \{ F_B^{t, i=0}, \cdots, F_B^{t, i=4}\right \}$ denotes the multi-scale features of the post-change image. Here, $i$ signifies the scale or level of the feature extracted by the DDPM decoder, with $i=4$ representing features at the lowest spatial resolution $\frac{h}{2^i} \times \frac{w}{2^i} = \frac{h}{16} \times \frac{w}{16}$. Similarly, $i=0$ corresponds to the highest spatial resolution, which matches the input image resolution $h \times w$.

We also obtain features corresponding to multiple noisy versions of the pre-change and post-change images. In our experiments, we demonstrate that combining feature representations from multiple noisy images can significantly enhance the change detection performance. Let the $n$ time-steps used to extract features be denoted as $\{ t_0, \cdots, t_n\}$. Hence, the multi-scale,  multi-timestep features of the pre-change image are given by:
\begin{equation}
    \left \{ F_A^{t, i=0}, \cdots, F_A^{t, i=4}\right \}_{t=t_0}^{t=t_n} = \left. f_{DDPM}(I_A^t) \right \vert_{t=t_0}^{t=t_n},
\end{equation}
and similarly for the post-change image:
\begin{equation}
    \left \{ F_B^{t, i=0}, \cdots, F_B^{t, i=4}\right \}_{t=t_0}^{t=t_n} = \left. f_{DDPM}(I_B^t) \right \vert_{t=t_0}^{t=t_n}.
\end{equation}
These multi-scale, multi-timestep features are then used to fine-tune a hierarchical change detection classifier.

Figure \ref{fig:feats} illustrates these multi-level multi-timestep feature representations for a given input image. These representations highlight DDPM's ability to extract hierarchical features (i.e., semantics) from input images. Features at larger scales, such as $h \times w$ (i.e., $i=0$) and $\frac{h}{2} \times \frac{w}{2}$ (i.e., $i=1$), capture simpler patterns like edges and gradients, while those at $\frac{h}{4} \times \frac{h}{4}$ (i.e., $i=2$), $\frac{h}{8} \times \frac{h}{8}$ (i.e., $i=3$), and $\frac{h}{16} \times \frac{h}{16}$ (i.e., $i=4$) scales represent more abstract features.

Unlike other deep learning architectures, DDPM enables the acquisition of diverse augmented feature representations by adjusting the amount of noise added to the input. These augmented representations can significantly enhance the performance of a change detection classifier, particularly in scenarios with limited pixel-level change labels. In our ablation study, we demonstrate that integrating multi-level features corresponding to various noisy image levels substantially improves the change detection performance.

\subsection{Fine-Tuning for Change Detection}
\label{sec:method-finetuning}
We proceed by fine-tuning a lightweight hierarchical change decoder and change classifier on the multi-scale, multi-timestep feature representations of pre-change and post-change images, as depicted in Figure \ref{fig:method}-III. The hierarchical change decoder, denoted as $f_d(\cdot)$, and the change classifier, denoted as $f_{cd}(\cdot)$, are fine-tuned using supervised change labels.

The hierarchical change decoder comprises of five change decoder blocks, denoted as $f_d^i(\cdot)$, responsible for extracting change features at each spatial scale $(i)$ from the multi-timestep $(t)$ features. At each scale, the difference feature representations are computed using the multi-timestep features of pre-change and post-change images and the difference features from the previous scale, as illustrated in Figure \ref{fig:method}-III. This process is executed from the lowest spatial scale ($i=4$) to the highest spatial scale ($i=0$). Subsequently, a simple two-layer change classifier is employed to obtain the final change predictions. The weights of the hierarchical change decoder and the classifier are updated using the Cross-Entropy (CE) loss during fine-tuning.

The hierarchical change block at $i+1$-th scale is depicted in Figure \ref{fig:method}-III. The feature representations corresponding to different noise levels $t=t_0,\cdots,t_n$ at a given scale $i+1$ are concatenated separately for pre-change and post-change image feature representations:
\begin{align}
    \overline{F}_{A}^{i+1} &= \texttt{CAT}\left(\left[F^{t=t_0,i+1}_{A}, \cdots, F^{t=t_n,i+1}_{A} \right],\hspace{2pt} \texttt{dim=1}\right),\\
    \overline{F}^{i+1}_{B} &= \texttt{CAT}\left(\left[F^{t=t_0,i+1}_{B}, \cdots, F^{t=t_n,i+1}_{B}\right],\hspace{2pt}  \texttt{dim=1}\right).
\end{align}
Here, $\texttt{CAT}$ denotes feature concatenation, considering the batch dimension as the $0-$th dimension. These concatenated feature representations undergo a series of two 2D convolutional layers. The first convolutional layer reduces the embedding dimension to the size of the embedding dimension before concatenation, $d^{i}$, while the second layer outputs feature representations of size $d^{i}$. This sequence acts as a dimensionality reduction layer, bringing down the embedding dimension from $n \times d^i$ to $d^i$.

Following this, the absolute difference between the feature representations $\overline{F}^{i+1}_{a}$ and $\overline{F}^{i+1}_{b}$ is computed since the objective is to detect changes in the input images:
\begin{equation}
    F_D^{i+1} = \left| \overline{F}_A^{i+1} - \overline{F}_B^{i+1} \right|_1,
\end{equation}
where $F_D^{i}$ signifies the difference feature representation at spatial scale $i+1$. Next, the difference feature representations from the current $F_D^{i+1}$ and previous scale (upsampled) $\widetilde{F}_D^{i}$ are combined by summation:
\begin{equation}
    \overline{F}_D^{i+1} = F_D^{i+1} + \texttt{Upsample}\left(\widetilde{F}_D^{i}, \texttt{factor}=2\right),
\end{equation}
with $\texttt{Upsample}(\cdot, \texttt{factor}=2)$ representing bi-linear upsampling by a factor of $2$. Subsequently, concurrent spatial-squeeze channel-excitation (denoted by $\texttt{cSE}$)  and channel-squeeze spatial-excitation module (denoted by $\texttt{sSE}$) are employed to enhance channel and spatial re-calibration of the difference feature representations at each scale. The $\texttt{cSE}$ operates by aggregating spatial information across the difference features and compressing it into channel-wise descriptor. This descriptor is used to re-weight channel importance, allowing dynamic re-calibration of difference features in each channel. On the other hand, $\texttt{sSE}$ compresses channel-wise information into spatial descriptors, which are then used to capture spatial dependencies across the channels. By adapting the spatial attention across channels, this module facilitates the network in attending to important spatial locations, enhancing the model's capability to capture intricate spatial patterns in the difference features. Mathematically, this process can be represented as:
\begin{equation}
    \widetilde{F}_D^{i+1}= \texttt{cSE}\left( \overline{F}_D^{i+1}\right)+\texttt{sSE}\left( \overline{F}_D^{i+1}\right),
\end{equation}
where $\widetilde{F}_D^{i+1}$ represents the output from the $i+1$-th hierarchical change decoder block.

This hierarchical change feature extraction continues until the last hierarchical change decoder block, i.e., $i=0$, obtaining the final difference feature representation $\widetilde{F}_D^{0}$:
\begin{align}
    \widetilde{F}_D^{4} &= f_d^4\left( 0, \{F_A^{t,i=4}\}_{t_0}^{t_n}, \{F_A^{t,i=4}\}_{t_0}^{t_n}\right), \\
    \widetilde{F}_D^{3} &= f_d^3\left( \widetilde{F}_D^{4}, \{F_A^{t,i=3}\}_{t_0}^{t_n}, \{F_A^{t,i=3}\}_{t_0}^{t_n}\right),\ \\ 
    &\vdots \nonumber \\
    \widetilde{F}_D^{0} &= f_d^0\left( \widetilde{F}_D^{1}, \{F_A^{t,i=0}\}_{t_0}^{t_n}, \{F_A^{t,i=0}\}_{t_0}^{t_n}\right).
\end{align}

Finally, the final change feature representations are processed through the change classifier, denoted by $f_{cls}(\cdot)$, to obtain the final change probability map $P_{cd}$:
\begin{equation}
    P_{cd} = f_{cls} \left( \widetilde{F}_D^{0} \right).
\end{equation}

The predicted change map $P_{cd}$ is compared with the ground truth change map $Y$ using the Binary Cross-Entropy loss to update the weights of the hierarchical change decoders and the change classifier during fine-tuning:
\begin{equation}
    \mathcal{L} = \texttt{CE} (P_{ce}, Y),
\end{equation}
where $\texttt{CE}$ denotes the pixel-wise binary Cross-Entropy loss.

\section{Experimental Setup}
\label{sec:experiments}

\subsection{Pre-training DDPM} 
We pre-train a state-of-the-art, pixel-space, unconditional U-Net-like diffusion model~\cite{saharia2021image} on remote sensing images collected from the Google Earth Engine without any human supervision. The pre-training is conducted on our DDPM using remote sensing images of resolution $256\times256$. Our DDPM model comprises of two convolutional residual blocks per resolution level and self-attention blocks at the $16 \times 16$ resolution, positioned between the convolutional blocks. Group normalization with a group size of $32$ is employed for normalization in both residual blocks and self-attention blocks, while a dropout rate of $0.2$ is utilized in residual blocks. All self-attention blocks consist of one attention head. Diffusion time $t$ is specified by adding the sinusoidal position embedding into each residual block. Our U-Net encoder includes five spatial scales (resolution levels). We set the base channel dimension to $128$ and define the channel multiplier as $\{1, 2, 4, 8, 8\}$. A cosine $\beta_t$ schedule is chosen, setting $T = 2000$ without a sweep, and implementing a linear schedule from $\beta_1 = 10^{-6}$ to $\beta_t = 0.01$. Following the practices in \cite{saharia2021image, ho2020denoising}, we utilize the $\texttt{Adam}$ optimizer with a linear warm-up schedule over $10,000$ training steps, followed by a fixed learning rate of $1 \times 10^{-5}$. The batch size is set to 8, and no exponential moving average (EMA) is applied to model parameters.

\subsection{Fine-tuning for change detection}
\textbf{Datasets:} Fine-tuning for change detection is conducted on four publicly available datasets: LEVIR-CD~\cite{LEVIR}, WHU-CD~\cite{WHU-CD}, DSIFN-CD~\cite{DSIFN}, and CDD~\cite{CDD}.

The LEVIR-CD~\cite{LEVIR} dataset comprises 637 pairs of high-resolution remote sensing images with a spatial size of $1024 \times 1024$ and a spatial resolution of $0.5 m$, collected from Google Earth. It provides a total of 31,333 change labels for building instances, covering both building appearances and disappearances. The dataset includes official splits for train, val, and test, with 44,564, 564, and 128 pairs, respectively. Each $1024 \times 1024$ image is pre-processed into non-overlapping $256 \times 256$ patches for train, val, and test sets.

The WHU-CD~\cite{WHU-CD} dataset consists of paired aerial images acquired in 2012 and 2016, covering an area of $20.5 km^2$, containing 12,796 and 16,077 building instances, respectively. The spatial size of each image is $15,354 \times 32,507$ pixels, with a spatial resolution of $0.2 m$. The dataset represents an area affected by a 6.3-magnitude earthquake in February 2011, resulting in numerous changes, including rebuilt buildings and new constructions. Semantic labels for building changes are available. Each image was divided into non-overlapping $256 \times 256$ patches for train, val, and test sets.

The DSIFN-CD~\cite{DSIFN} dataset comprises of six bi-temporal high-resolution images from six cities in China, clipped into 394 sub-image pairs of sizes $512 \times 512$. After augmentation, it contains 3940 bi-temporal image pairs. The training set includes 3600 image pairs, the validation set includes 340 image pairs, and the test set includes 48 image pairs. Each image was preprocessed into non-overlapping $256 \times 256$ patches for train, val, and test sets.

The CDD~\cite{CDD} dataset consists of season-varying remote sensing images obtained from Google Earth. It contains 7 pairs of season-varying images with a resolution of $4725 \times 2700$ pixels for manual ground truth creation and 4 pairs with minimal changes and a resolution of $1900 \times 1000$ pixels for additional manual object inclusion. The dataset features objects of varying sizes (e.g., cars to large construction structures) and seasonal changes in natural objects (e.g., from a single tree to a wide forest area). The final dataset contains  $16000$ image pairs of size $256 \times 256$: $10000$ for training, $3000$ for validation, and $3000$ for testing.

\textbf{Fine-tuning:} We optimize the parameters of the hierarchical change decoder and the change classifier using the Binary Cross-Entropy (CE) loss and the $\texttt{AdamW}$ optimizer. The parameters of the pre-trained DDPM remain frozen during the fine-tuning process. Our initial learning rate is set to $1 \times 10^{-4}$, linearly decaying to zero over $120$ epochs. The evaluation metrics are reported on the $\texttt{test}$-set.

\subsection{Performance metrics}
In order to measure the change detection performance, we use $F1$ and Intersection over Union (IoU) scores with regards to the \textit{change class} as the primary quantitative indices~\cite{bandara2022revisiting, bandara2022transformer, chen2021remote}. Additionally, we report the overall accuracy (OA) to get a global quality of change predictions.

\renewcommand{\tabcolsep}{3pt}
\begin{table*}[!tb]
    \centering
    \caption{\textbf{The average quantitative change detection results on the LEVIR-CD, WHU-CD, DSIFN-CD, and CDD \texttt{test}-sets.} Our DDPM-CD achieves the best results in F1, IoU, and OA metrics (highlited in red color) on all four datasets. $\uparrow$ indicates higher the value better the change detection performance. ``-'' indicates not reported or not available to us. (IN1k) indicates pre-training process is initialized with the ImageNet pre-trained weights. IN1k, IBSD, and GE refers to ImageNet1k~\cite{krizhevsky2012imagenet}, Inria Building Segmentation Dataset~\cite{inria}, and Google Earth. }
    \begin{tabular}{l|p{.05\linewidth}|p{.12\linewidth}|p{.05\linewidth}|ccccccccccccccc}
		\toprule
		\multirow{2}{*}{Method} & \multirow{2}{*}{\parbox{\linewidth}{Pre-train ?}} & \multirow{2}{*}{\parbox{\linewidth}{\centering Extra data}} & \multirow{2}{*}{\parbox{\linewidth}{Extra labels}} & \multicolumn{3}{c}{LEVIR-CD~\cite{LEVIR}} && 
		\multicolumn{3}{c}{WHU-CD~\cite{WHU-CD}}  &&
		\multicolumn{3}{c}{DSIFN-CD~\cite{DSIFN}} &&
		\multicolumn{3}{c}{CDD~\cite{CDD}}        \\ 
		\cmidrule{5-7} \cmidrule{9-11} \cmidrule{13-15} \cmidrule{17-19}
		&&&&
        {F1 $\uparrow$}&{IoU $\uparrow$}&{OA $\uparrow$}&&
		{F1 $\uparrow$}&{IoU $\uparrow$}&{OA $\uparrow$}&&
		{F1 $\uparrow$}&{IoU $\uparrow$}&{OA $\uparrow$}&&
		{F1 $\uparrow$}&{IoU $\uparrow$}&{OA $\uparrow$}\\
		\midrule
		
        \rowcolor{gray}
        \multicolumn{19}{l}{\bf Methods trained from random initialization:}  \\
		FC-EF\cite{daudt2018fully} & \xmark & \xmark & \xmark &
                                    83.40 & 71.53 & 98.39&&
		                            76.73 & 62.24 & 98.31&&
		                            72.17 & 56.45 & 83.37&&
		                            66.93 & 50.30 & 93.28\\
		FC-Siam-diff\cite{daudt2018fully}& \xmark & \xmark & \xmark &
                              86.31 & 75.92 & 98.67&&
		                      78.95 & 65.23 & 98.48&&
		                      70.55 & 54.49 & 84.13&&
		                      70.61 & 54.57 & 94.33\\
		FC-Siam-conc\cite{daudt2018fully}& \xmark & \xmark & \xmark &
                          83.69 & 71.96 & 98.49&&
		                  79.85 & 66.46 & 98.50&&
		                  59.71 & 42.56 & 87.57&&
		                  75.11 & 60.14 & 94.95\\
        SNUNet\cite{fang2021snunet}& \xmark & \xmark & \xmark &
                          88.16 & 78.83 & 98.82&&
		                  83.50 & 71.67 & 98.71&&
		                  66.18 & 49.45 & 87.34&&
		                  83.89 & 72.11 & 96.23\\ 
        ChangeFormer\cite{bandara2022transformer}&\cmark &\xmark &\cmark &                   90.40 & 82.48 & 99.04 &&
		                  88.57 & 79.49 & 99.12 &&
		                  94.67 & 88.71 & 93.23 &&
		                  94.63 & 89.80 & 98.74 \\ \\
                  
        \rowcolor{gray}
        \multicolumn{19}{l}{\bf Methods using supervised pre-trained weights:}  \\
		DT-SCN\cite{liu2020building}& \cmark & IN1k & \cmark &
                          87.67 & 78.05 & 98.77&&
		                  91.43 & 84.21 & 99.35&&
		                  70.58 & 54.53 & 82.87&&
		                  92.09 & 85.34 & 98.16\\
		STANet\cite{chen2020spatial}& \cmark & IN1k & \cmark &
                          87.26 & 77.40 & 98.66&&
		                  82.32 & 69.95 & 98.52&&
		                  64.56 & 47.66 & 88.49&&
		                  84.12 & 72.22 & 96.13\\
		IFNet\cite{zhang2020deeply} & \cmark & IN1k & \cmark &
                          88.13 & 78.77 & 98.87&&
		                  83.40 & 71.52 & 98.83&&
		                  60.10 & 42.96 & 87.83&&
		                  84.00 & 71.91 & 96.03\\
        BIT\cite{chen2021remote}& \cmark & IN1k & \cmark &
                          89.31 & 80.68 & 98.92&&
		                  90.53 & 83.39 & 99.34&&
		                  87.61 & 77.96 & 92.30&&
		                  88.90 & 80.01 & 97.47\\ \\

        \rowcolor{gray}
        \multicolumn{19}{l}{\bf Methods fine-tuned using self-supervised pre-trained weights on remote sensing images:}  \\
        SimSiam\cite{chen2020simple} & \cmark & (IN1k), IBSD, GE & \xmark &
                        88.00 & 78.57 & - &&
                        84.69 & 73.45 & - &&
                        -    & -    & - &&
                        -    & -    & - \\
        MoCo-v2\cite{nichol2021improved} & \cmark & (IN1k), IBSD, GE & \xmark &
                        88.21 & 78.90 & - &&
                        87.90 & 78.41 & - &&
                        -    & -    & - &&
                        -    & -    & - \\
        DenseCL\cite{wang2021dense} & \cmark & (IN1k), IBSD, GE & \xmark &
                        86.69 & 76.51 & - &&
                        86.94 & 76.90 & - &&
                        -    & -    & - &&
                        -    & -    & - \\
        CMC\cite{cmc}     & \cmark & (IN1k), IBSD, GE & \xmark &
                        88.61 & 79.55 & - &&
                        87.66 & 78.03 & - &&
                        -    & -    & - &&
                        -    & -    & - \\
        SeCo\cite{manas2021seasonal}& \cmark & (IN1k), IBSD, GE & \xmark & 
                        88.27 & 79.01 & - &&
                        88.09 & 78.72 & - &&
                        -    & -    & - &&
                        -    & -    & - \\
        SaDL-CD\cite{manas2021seasonal}& \cmark & (IN1k), IBSD, GE & \cmark & 
                        88.74 & 79.75 & - &&
                        89.98 & 81.78 & - &&
                        -    & -    & - &&
                        -    & -    & - \\
        \midrule
		DDPM-CD (ours)  & \cmark & Google Earth & \xmark &
                        \textcolor{red}{\bf 90.91} & \textcolor{red}{\bf 83.35} &  \textcolor{red}{\bf 99.09} &&
		                \textcolor{red}{\bf92.65}  & \textcolor{red}{\bf86.31} & \textcolor{red}{\bf99.42} &&
		                \textcolor{red}{\bf 96.65} & \textcolor{red}{\bf91.28} & \textcolor{red}{\bf97.09}&&
		                \textcolor{red}{\bf 95.62} & \textcolor{red}{\bf91.62} & \textcolor{red}{\bf98.98}\\
		\bottomrule
	\end{tabular}   
    \label{tab:quant_res}
\end{table*}

\section{Results and Discussion}
In this section, we compare the proposed DDPM-CD method with several SOTA methods: Fully-Convolutional Early-Fusion (FC-EF)~\cite{daudt2018fully}, Fully-Convolutional Siamese-Difference (FC-Saim-diff)~\cite{daudt2018fully}, Fully-Convolutional Siamese-Concatenation (FC-Siam-conc)~\cite{daudt2018fully}, Dual-Task Constrained Siamese Network (DT-SCN)~\cite{liu2020building}, Spatial-Temporal Attention Network (STA-Net)~\cite{chen2020spatial}, Densely Connected Siamese Network (SNU-Net)~~\cite{fang2021snunet}, Bi-Temporal Image Transformer (BIT)~\cite{chen2021remote}, and Transformer-based Siamese Network (ChangeFormer)~\cite{bandara2022transformer}. Notably, BIT and ChangeFormer are the latest SOTA methods employing non-local-attention networks, specifically transformers, for remote sensing change detection.

Among these methods, FC-EF, FC-SD, and FC-SC start training from random initialization without utilizing pre-trained weights. DT-SCN uses a pre-trained ResNet with Squeeze-and-Excitation Networks (SE-ResNet)~\cite{hu2018squeeze}\footnote{\url{https://github.com/hujie-frank/SENet}} on ImageNet-1k (with supervised labels) to obtain initial feature representations. Similarly, STANet utilizes a pre-trained ResNet-18 on ImageNet (with supervised labels) for feature extraction. IFNet also uses VGG16 pre-trained on ImageNet (with supervised labels) to extract features from bi-temporal images. In contrast, SNUNet~\cite{fang2021snunet}\footnote{\url{https://github.com/likyoo/Siam-NestedUNet/}} begins training from random initialization without pre-trained weights.

In recent transformer-based approaches, BIT~\cite{chen2021remote}\footnote{\url{https://github.com/justchenhao/BIT_CD}} uses pre-trained ResNet-50 on ImageNet1k (with supervised labels) as the feature extractor to obtain tokens that feed into the transformer encoder. Conversely, in the latest fully transformer-based method, ChangeFormer~\cite{bandara2022transformer}\footnote{\url{https://github.com/wgcban/ChangeFormer}}, hierarchical transformer blocks are initialized with pre-trained weights on ADE20k for supervised segmentation tasks before fine-tuning for change detection.

We provide a summary of the pre-training details related to these methods along with the quantitative change detection results in Table \ref{tab:quant_res} for proper comparison between the methods. The table includes details on whether pre-trained weights are used in the feature extractor (in second column), weather the pre-training involves additional data apart from the fine-tuning dataset (in third column), and if the pre-training requires supervised labels (in fourth column).

\subsection{Quantitative change detection results}
Table \ref{tab:quant_res} presents a quantitative comparison of the change detection performance between the proposed method and the state-of-the-art methods, evaluated based on the F1, IoU, and OA scores. The reported numbers for the proposed DDPM-CD represent the best (default) configuration, where we utilize multi-timestep features at timesteps $t$ equal to 50, 100, and 400 (refer to Sec. \ref{sec:ablation} for further details on how performance varies based on the timestep used to obtain features). Considering the default configuration of DDPM-CD, a significant performance improvement over existing SOTA change detection methods based on random initialization, supervised pre-trained backbones, and self-supervised pre-trained backbones. When considering methods that depend on random initialization, the improvement from our method is significant, highlighting the importance of having pre-trained weights. Hence, in most of the recently proposed methods, pre-trained weights on another dataset are utilized to initialize the change detection training. Since ImageNet (IN1k) supervised pre-trained weights are widely adopted and publically available, it is utilized in change detection as well, as seen in DT-SCN, STANet, IFNet, and BIT. However, natural images and remote sensing images are fundamentally different, and ImageNet pre-trained weights are not ideally suitable for remote sensing applications, resulting in sub-optimal results. Later, many self-supervised approaches have been developed such as in SimSiam, MoCo-v2, DenseCL, CMC, SeCo, and SaDL-CD and utilized for change detection. It is important to note that most of these methods still initialize the pre-training with ImageNet pre-trained weights to speed up their pre-training. We can see that our proposed DDPM-CD considerably outperforms these recent change detection methods based on self-supervised pre-trained weights, empirical demonstrating the effectiveness of diffusion models in extracting crucial semantics for change detection.
%
Moreover, it is important to note that DDPM-CD solely utilizes off-the-shelf remote sensing images from the Google Earth Engine for pre-training, eliminating the necessity of large classification datasets for supervised pre-training of the feature extractor. This stands in stark contrast with the previous SOTA methods like BIT and ChangeFormer, which rely on pre-training backbones with supervised large-scale classification datasets such as ImageNet and ADE20k. The substantial improvement in change detection performance compared to these state-of-the-art methods further validates the capability of pre-trained diffusion models in generating meaningful representations that are valuable for change detection.

\begin{figure}[htb!]
    \begin{subfigure}[t]{.2\linewidth}
        \centering
        \includegraphics[width=.99\textwidth]{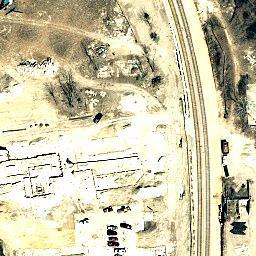}
        \caption{}
    \end{subfigure}%
    \begin{subfigure}[t]{.2\linewidth}
        \centering
        \includegraphics[width=.99\textwidth]{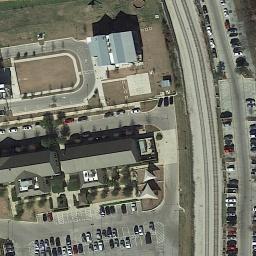}
        \caption{}
    \end{subfigure}%
    \begin{subfigure}[t]{.2\linewidth}
        \centering
        \includegraphics[width=.99\textwidth]{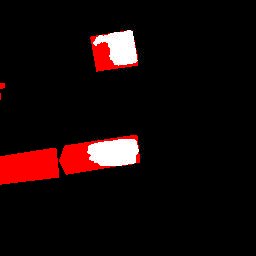}
        \caption{}
    \end{subfigure}%
    \begin{subfigure}[t]{.2\linewidth}
        \centering
        \includegraphics[width=.99\textwidth]{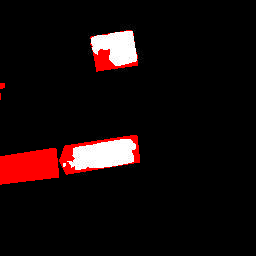}
        \caption{}
    \end{subfigure}%
    \begin{subfigure}[t]{.2\linewidth}
        \centering
        \includegraphics[width=.99\textwidth]{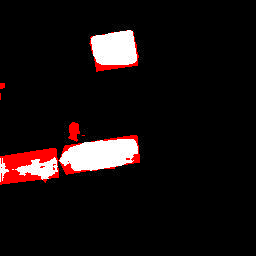}
        \caption{}
    \end{subfigure}%
    \\
    \begin{subfigure}[t]{.2\linewidth}
        \centering
        \includegraphics[width=.99\textwidth]{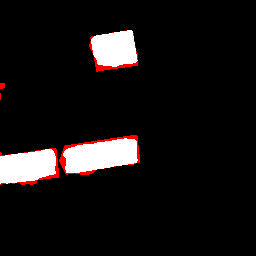}
        \caption{}
    \end{subfigure}%
    \begin{subfigure}[t]{.2\linewidth}
        \centering
        \includegraphics[width=.99\textwidth]{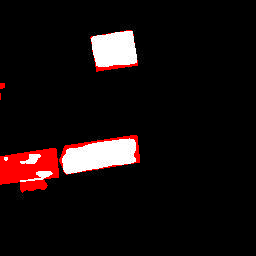}
        \caption{}
    \end{subfigure}%
    \begin{subfigure}[t]{.2\linewidth}
        \centering
        \includegraphics[width=.99\textwidth]{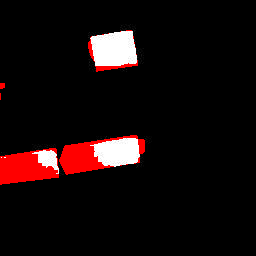}
        \caption{}
    \end{subfigure}%
    \begin{subfigure}[t]{.2\linewidth}
        \centering
        \includegraphics[width=.99\textwidth]{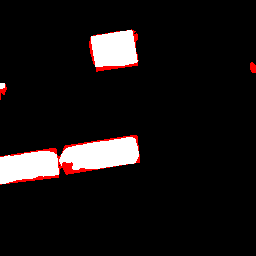}
        \caption{}
    \end{subfigure}%
    \begin{subfigure}[t]{.2\linewidth}
        \centering
        \includegraphics[width=.99\textwidth]{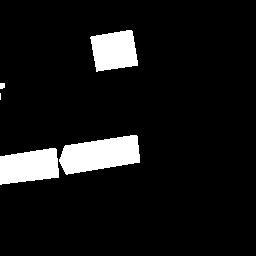}
        \caption{}
    \end{subfigure}%
    
    \vspace{1mm}
    \centering \text{Case-I}\\
    \vspace{3mm}
    \setcounter{subfigure}{0}
    
    \begin{subfigure}[t]{.2\linewidth}
        \centering
        \begin{tikzpicture}
            \node[anchor=south west,inner sep=0] at (0,0) {\includegraphics[width=.99\textwidth]{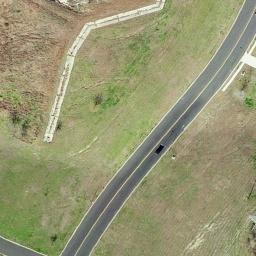}};
        \end{tikzpicture}
        \caption{}
    \end{subfigure}%
    \begin{subfigure}[t]{.2\linewidth}
        \centering
        \begin{tikzpicture}
            \node[anchor=south west,inner sep=0] at (0,0) {\includegraphics[width=.99\textwidth]{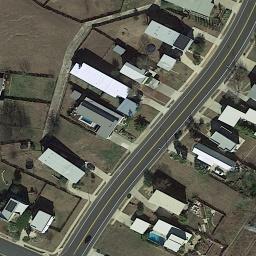}};
        \end{tikzpicture}
        \caption{}
    \end{subfigure}%
    \begin{subfigure}[t]{.2\linewidth}
        \centering
        \begin{tikzpicture}
            \node[anchor=south west,inner sep=0] at (0,0) {        \includegraphics[width=.99\textwidth]{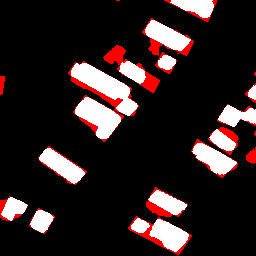}};
        \end{tikzpicture}
        \caption{}
        \label{fig:1c}
    \end{subfigure}%
    \begin{subfigure}[t]{.2\linewidth}
        \centering
        \begin{tikzpicture}
            \node[anchor=south west,inner sep=0] at (0,0) {        \includegraphics[width=.99\textwidth]{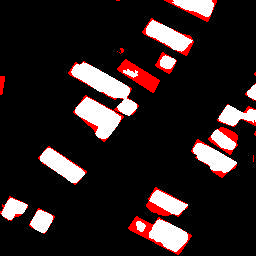}};
        \end{tikzpicture}
        \caption{}
    \end{subfigure}%
    \begin{subfigure}[t]{.2\linewidth}
        \centering
        \begin{tikzpicture}
            \node[anchor=south west,inner sep=0] at (0,0) {        \includegraphics[width=.99\textwidth]{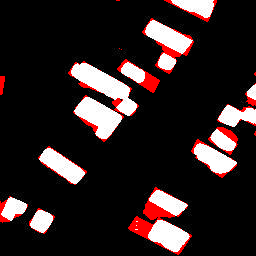}};
        \end{tikzpicture}
        \caption{}
    \end{subfigure}%
    \\
    \begin{subfigure}[t]{.2\linewidth}
        \centering
        \begin{tikzpicture}
            \node[anchor=south west,inner sep=0] at (0,0) {        \includegraphics[width=.99\textwidth]{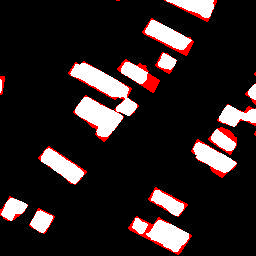}};
        \end{tikzpicture}
        \caption{}
    \end{subfigure}%
    \begin{subfigure}[t]{.2\linewidth}
        \centering
        \begin{tikzpicture}
            \node[anchor=south west,inner sep=0] at (0,0) {        \includegraphics[width=.99\textwidth]{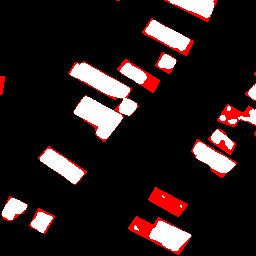}};
        \end{tikzpicture}
        \caption{}
    \end{subfigure}%
    \begin{subfigure}[t]{.2\linewidth}
        \centering
        \begin{tikzpicture}
            \node[anchor=south west,inner sep=0] at (0,0) {        \includegraphics[width=.99\textwidth]{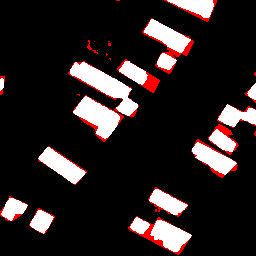}};
        \end{tikzpicture}
        \caption{}
    \end{subfigure}%
    \begin{subfigure}[t]{.2\linewidth}
        \centering
        \begin{tikzpicture}
            \node[anchor=south west,inner sep=0] at (0,0) {\includegraphics[width=.99\textwidth]{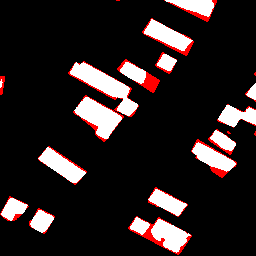}};
        \end{tikzpicture}
        \caption{}
    \end{subfigure}%
    \begin{subfigure}[t]{.2\linewidth}
        \centering
        \begin{tikzpicture}
            \node[anchor=south west,inner sep=0] at (0,0) {        \includegraphics[width=.99\textwidth]{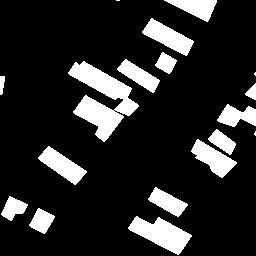}};
        \end{tikzpicture}
        \caption{}
    \end{subfigure}%
    \vspace{1mm}
    \centering \text{Case-II}\\
    
    \captionof{figure}{Comparison of different state-of-the-art change detection methods on \textbf{LEVIR-CD} dataset: \textbf{(a)} Pre-change image, \textbf{(b)} Post-change image, \textbf{(c)} FC-EF, \textbf{(d)} FC-Siam-diff, \textbf{(e)} FC-Siam-conc, \textbf{(f)} DT-SCN, \textbf{(g)} BIT, \textbf{(h)} ChangeFormer, \textbf{(i)} DDPM-CD (ours), and \textbf{(j)} Ground-truth. \textit{Note that true positives (change class) are indicated in white, true negatives (no-change class) are indicated in black, and false positives plus false negatives indicates in red.}}
    \label{fig:visual_levir}
\end{figure}
\begin{figure}[htb!]
    \begin{subfigure}[t]{.2\linewidth}
        \centering
        \includegraphics[width=.99\textwidth]{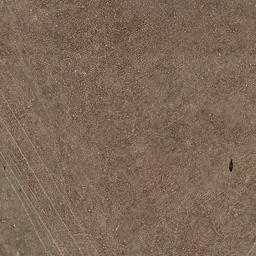}
        \caption{}
        
    \end{subfigure}%
    \begin{subfigure}[t]{.2\linewidth}
        \centering
        \includegraphics[width=.99\textwidth]{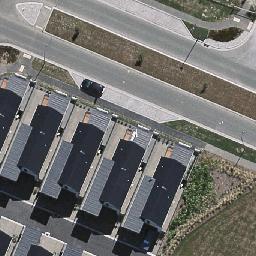}
        \caption{}
    \end{subfigure}%
    \begin{subfigure}[t]{.2\linewidth}
        \centering
        \includegraphics[width=.99\textwidth]{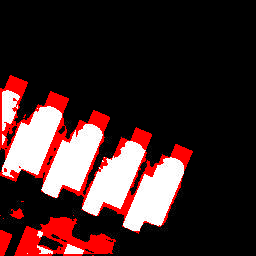}
        \caption{}
    \end{subfigure}%
    \begin{subfigure}[t]{.2\linewidth}
        \centering
        \includegraphics[width=.99\textwidth]{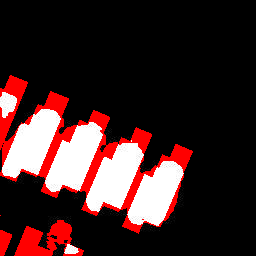}
        \caption{}
    \end{subfigure}%
    \begin{subfigure}[t]{.2\linewidth}
        \centering
        \includegraphics[width=.99\textwidth]{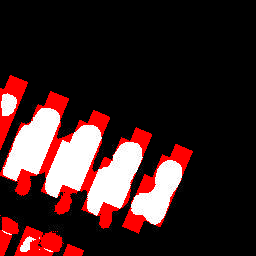}
        \caption{}
    \end{subfigure}%
    \\
    \begin{subfigure}[t]{.2\linewidth}
        \centering
        \includegraphics[width=.99\textwidth]{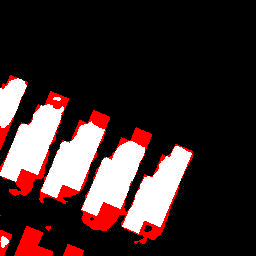}
        \caption{}
    \end{subfigure}%
    \begin{subfigure}[t]{.2\linewidth}
        \centering
        \includegraphics[width=.99\textwidth]{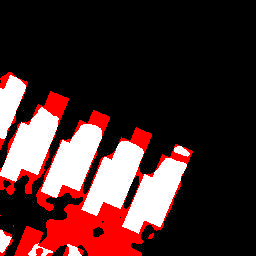}
        \caption{}
    \end{subfigure}%
    \begin{subfigure}[t]{.2\linewidth}
        \centering
        \includegraphics[width=.99\textwidth]{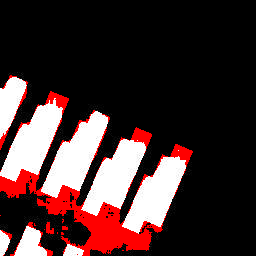}
        \caption{}
    \end{subfigure}%
    \begin{subfigure}[t]{.2\linewidth}
        \centering
        \includegraphics[width=.99\textwidth]{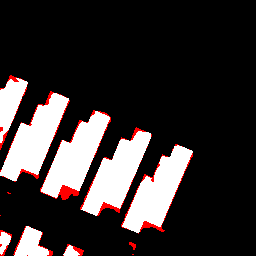}
        \caption{}
    \end{subfigure}%
    \begin{subfigure}[t]{.2\linewidth}
        \centering
        \includegraphics[width=.99\textwidth]{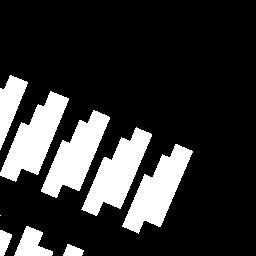}
        \caption{}
    \end{subfigure}%

   \vspace{1mm}
    \centering \text{Case-I}\\
    \vspace{3mm}
    \setcounter{subfigure}{0}
    
    \begin{subfigure}[t]{.2\linewidth}
        \centering
        \includegraphics[width=.99\textwidth]{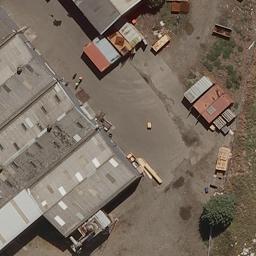}
        \caption{}
        \label{fig:1a}
    \end{subfigure}%
    \begin{subfigure}[t]{.2\linewidth}
        \centering
        \includegraphics[width=.99\textwidth]{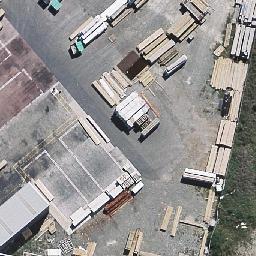}
        \caption{}
    \end{subfigure}%
    \begin{subfigure}[t]{.2\linewidth}
        \centering
        \includegraphics[width=.99\textwidth]{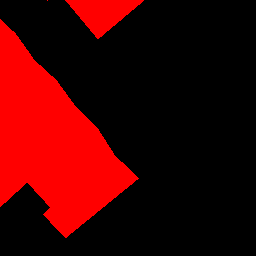}
        \caption{}
    \end{subfigure}%
    \begin{subfigure}[t]{.2\linewidth}
        \centering
        \includegraphics[width=.99\textwidth]{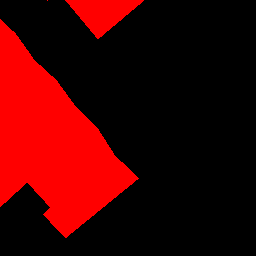}
        \caption{}
    \end{subfigure}%
    \begin{subfigure}[t]{.2\linewidth}
        \centering
        \includegraphics[width=.99\textwidth]{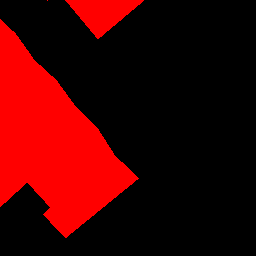}
        \caption{}
    \end{subfigure}%
    \\
    \begin{subfigure}[t]{.2\linewidth}
        \centering
        \includegraphics[width=.99\textwidth]{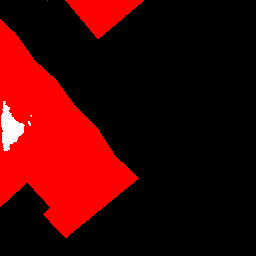}
        \caption{}
    \end{subfigure}%
    \begin{subfigure}[t]{.2\linewidth}
        \centering
        \includegraphics[width=.99\textwidth]{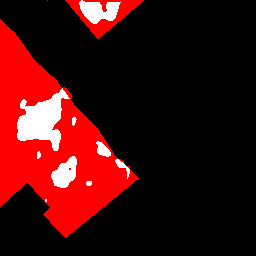}
        \caption{}
    \end{subfigure}%
    \begin{subfigure}[t]{.2\linewidth}
        \centering
        \includegraphics[width=.99\textwidth]{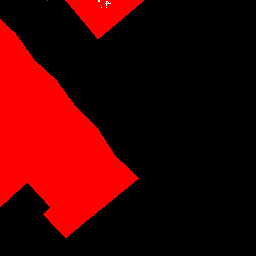}
        \caption{}
    \end{subfigure}%
    \begin{subfigure}[t]{.2\linewidth}
        \centering
        \includegraphics[width=.99\textwidth]{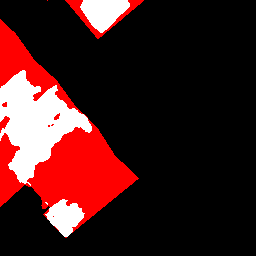}
        \caption{}
    \end{subfigure}%
    \begin{subfigure}[t]{.2\linewidth}
        \centering
        \includegraphics[width=.99\textwidth]{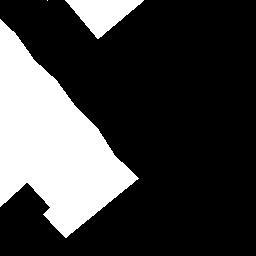}
        \caption{}
    \end{subfigure}%
    \vspace{1mm}
    \centering \text{Case-II}\\
    
    \captionof{figure}{Comparison of different state-of-the-art change detection methods on \textbf{WHU-CD} dataset: \textbf{(a)} Pre-change image, \textbf{(b)} Post-change image, \textbf{(c)} FC-EF, \textbf{(d)} FC-Siam-diff, \textbf{(e)} FC-Siam-conc, \textbf{(f)} DT-SCN, \textbf{(g)} BIT, \textbf{(h)} ChangeFormer, \textbf{(i)} DDPM-CD (ours), and \textbf{(j)} Ground-truth. \textit{Note that true positives (change class) are indicated in white, true negatives (no-change class) are indicated in black, and false positives plus false negatives indicates in red.}}
    \label{fig:visual_whu}
\end{figure}
\begin{figure}[htb!]
    \begin{subfigure}[t]{.2\linewidth}
        \centering
        \includegraphics[width=.99\textwidth]{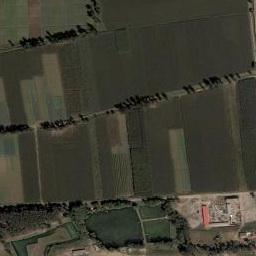}
        \caption{}
    \end{subfigure}%
    \begin{subfigure}[t]{.2\linewidth}
        \centering
        \includegraphics[width=.99\textwidth]{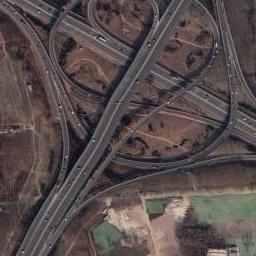}
        \caption{}
    \end{subfigure}%
    \begin{subfigure}[t]{.2\linewidth}
        \centering
        \includegraphics[width=.99\textwidth]{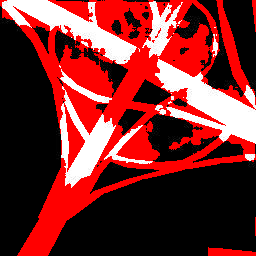}
        \caption{}
    \end{subfigure}%
    \begin{subfigure}[t]{.2\linewidth}
        \centering
        \includegraphics[width=.99\textwidth]{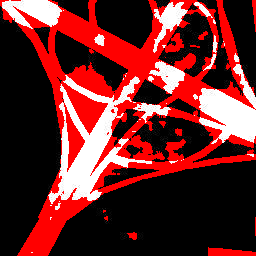}
        \caption{}
    \end{subfigure}%
    \begin{subfigure}[t]{.2\linewidth}
        \centering
        \includegraphics[width=.99\textwidth]{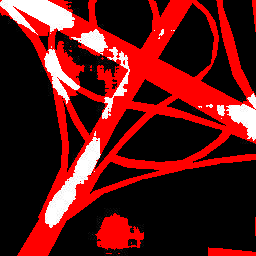}
        \caption{}
    \end{subfigure}%
    \\
    \begin{subfigure}[t]{.2\linewidth}
        \centering
        \includegraphics[width=.99\textwidth]{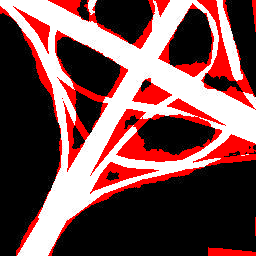}
        \caption{}
    \end{subfigure}%
    \begin{subfigure}[t]{.2\linewidth}
        \centering
        \includegraphics[width=.99\textwidth]{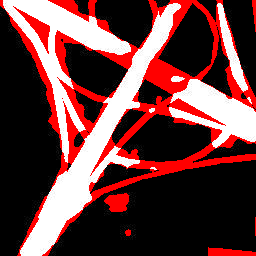}
        \caption{}
    \end{subfigure}%
    \begin{subfigure}[t]{.2\linewidth}
        \centering
        \includegraphics[width=.99\textwidth]{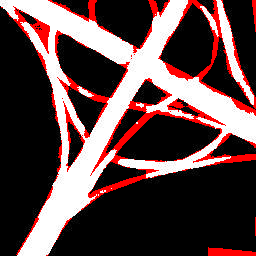}
        \caption{}
    \end{subfigure}%
    \begin{subfigure}[t]{.2\linewidth}
        \centering
        \includegraphics[width=.99\textwidth]{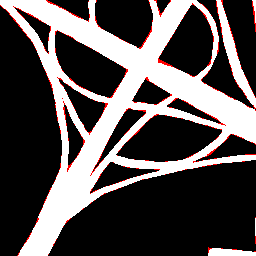}
        \caption{}
    \end{subfigure}%
    \begin{subfigure}[t]{.2\linewidth}
        \centering
        \includegraphics[width=.99\textwidth]{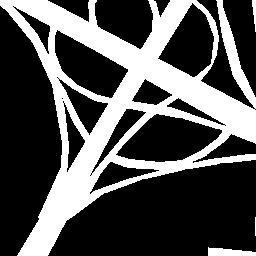}
        \caption{}
    \end{subfigure}%

    \vspace{1mm}
    \centering \text{Case-I}\\
    \vspace{3mm}
    \setcounter{subfigure}{0}
    
    \begin{subfigure}[t]{.2\linewidth}
        \centering
        \includegraphics[width=.99\textwidth]{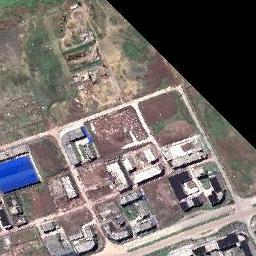}
        \caption{}
    \end{subfigure}%
    \begin{subfigure}[t]{.2\linewidth}
        \centering
        \includegraphics[width=.99\textwidth]{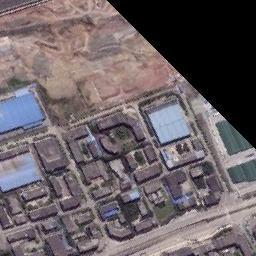}
        \caption{}
    \end{subfigure}%
    \begin{subfigure}[t]{.2\linewidth}
        \centering
        \includegraphics[width=.99\textwidth]{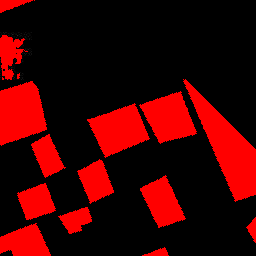}
        \caption{}
    \end{subfigure}%
    \begin{subfigure}[t]{.2\linewidth}
        \centering
        \includegraphics[width=.99\textwidth]{imgs/colored_pred/DSIFN_unet_770}
        \caption{}
    \end{subfigure}%
    \begin{subfigure}[t]{.2\linewidth}
        \centering
        \includegraphics[width=.99\textwidth]{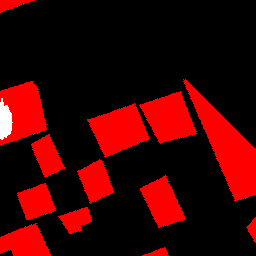}
        \caption{}
    \end{subfigure}%
    \\
    \begin{subfigure}[t]{.2\linewidth}
        \centering
        \includegraphics[width=.99\textwidth]{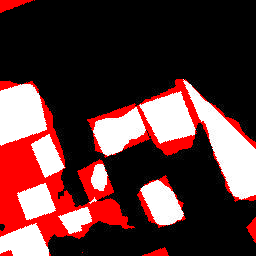}
        \caption{}
    \end{subfigure}%
    \begin{subfigure}[t]{.2\linewidth}
        \centering
        \includegraphics[width=.99\textwidth]{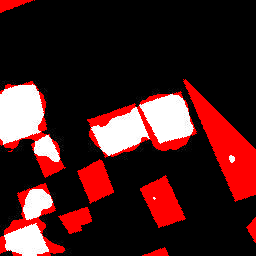}
        \caption{}
    \end{subfigure}%
    \begin{subfigure}[t]{.2\linewidth}
        \centering
        \includegraphics[width=.99\textwidth]{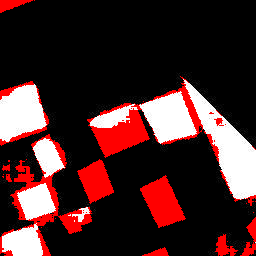}
        \caption{}
    \end{subfigure}%
    \begin{subfigure}[t]{.2\linewidth}
        \centering
        \includegraphics[width=.99\textwidth]{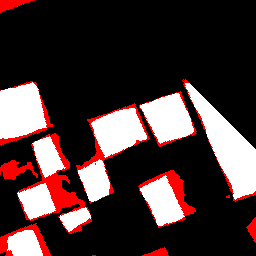}
        \caption{}
    \end{subfigure}%
    \begin{subfigure}[t]{.2\linewidth}
        \centering
        \includegraphics[width=.99\textwidth]{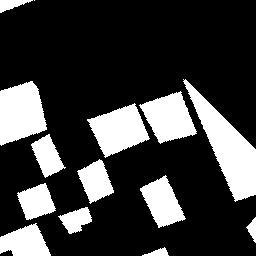}
        \caption{}
    \end{subfigure}%
    \vspace{1mm}
    \centering \text{Case-II}\\
    
    \captionof{figure}{Comparison of different state-of-the-art change detection methods on \textbf{DSIFN-CD} dataset: \textbf{(a)} Pre-change image, \textbf{(b)} Post-change image, \textbf{(c)} FC-EF, \textbf{(d)} FC-Siam-diff, \textbf{(e)} FC-Siam-conc, \textbf{(f)} DT-SCN, \textbf{(g)} BIT, \textbf{(h)} ChangeFormer, \textbf{(i)} DDPM-CD (ours), and \textbf{(j)} Ground-truth. \textit{Note that true positives (change class) are indicated in white, true negatives (no-change class) are indicated in black, and false positives plus false negatives indicates in red.}}
    \label{fig:visual_dsifn}
\end{figure}
\begin{figure}[htb!]
    \begin{subfigure}[t]{.2\linewidth}
        \centering
        \includegraphics[width=.99\textwidth]{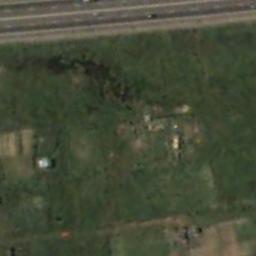}
        \caption{}
    \end{subfigure}%
    \begin{subfigure}[t]{.2\linewidth}
        \centering
        \includegraphics[width=.99\textwidth]{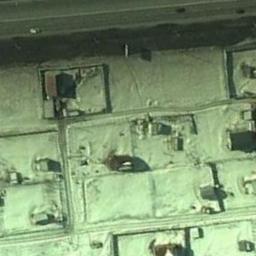}
        \caption{}
    \end{subfigure}%
    \begin{subfigure}[t]{.2\linewidth}
        \centering
        \includegraphics[width=.99\textwidth]{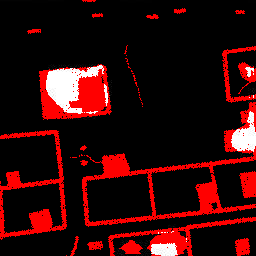}
        \caption{}
    \end{subfigure}%
    \begin{subfigure}[t]{.2\linewidth}
        \centering
        \includegraphics[width=.99\textwidth]{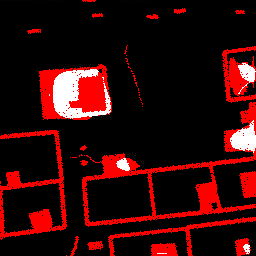}
        \caption{}
    \end{subfigure}%
    \begin{subfigure}[t]{.2\linewidth}
        \centering
        \includegraphics[width=.99\textwidth]{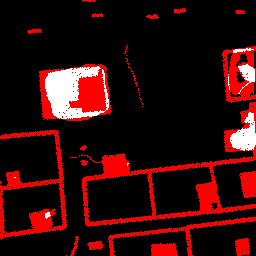}
        \caption{}
    \end{subfigure}%
    \\
    \begin{subfigure}[t]{.2\linewidth}
        \centering
        \includegraphics[width=.99\textwidth]{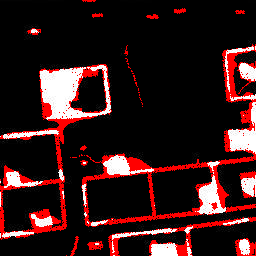}
        \caption{}
    \end{subfigure}%
    \begin{subfigure}[t]{.2\linewidth}
        \centering
        \includegraphics[width=.99\textwidth]{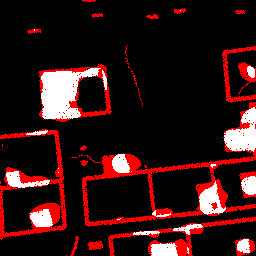}
        \caption{}
    \end{subfigure}%
    \begin{subfigure}[t]{.2\linewidth}
        \centering
        \includegraphics[width=.99\textwidth]{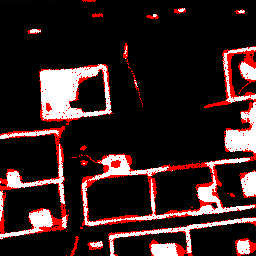}
        \caption{}
    \end{subfigure}%
    \begin{subfigure}[t]{.2\linewidth}
        \centering
        \includegraphics[width=.99\textwidth]{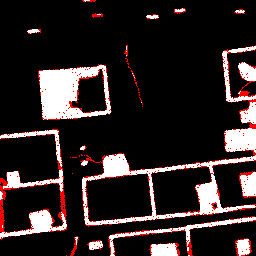}
        \caption{}
    \end{subfigure}%
    \begin{subfigure}[t]{.2\linewidth}
        \centering
        \includegraphics[width=.99\textwidth]{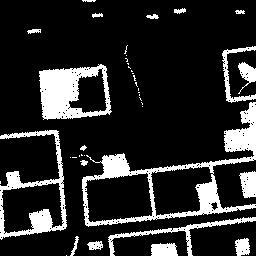}
        \caption{}
    \end{subfigure}%

    \vspace{1mm}
    \centering \text{Case-I}\\
    \vspace{3mm}
    \setcounter{subfigure}{0}
    
    \begin{subfigure}[t]{.2\linewidth}
        \centering
        \includegraphics[width=.99\textwidth]{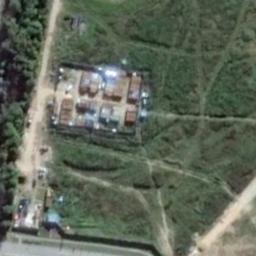}
        \caption{}
    \end{subfigure}%
    \begin{subfigure}[t]{.2\linewidth}
        \centering
        \includegraphics[width=.99\textwidth]{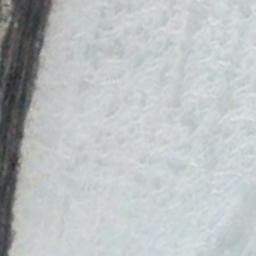}
        \caption{}
    \end{subfigure}%
    \begin{subfigure}[t]{.2\linewidth}
        \centering
        \includegraphics[width=.99\textwidth]{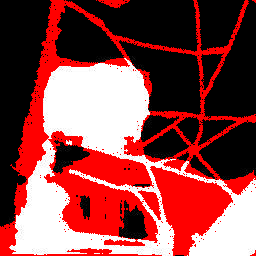}
        \caption{}
    \end{subfigure}%
    \begin{subfigure}[t]{.2\linewidth}
        \centering
        \includegraphics[width=.99\textwidth]{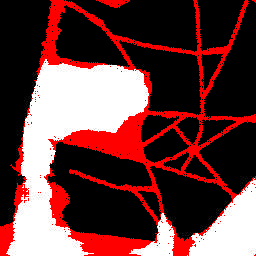}
        \caption{}
    \end{subfigure}%
    \begin{subfigure}[t]{.2\linewidth}
        \centering
        \includegraphics[width=.99\textwidth]{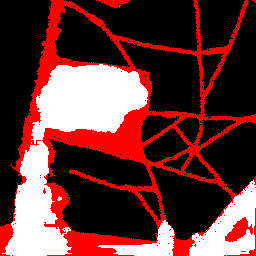}
        \caption{}
    \end{subfigure}%
    \\
    \begin{subfigure}[t]{.2\linewidth}
        \centering
        \includegraphics[width=.99\textwidth]{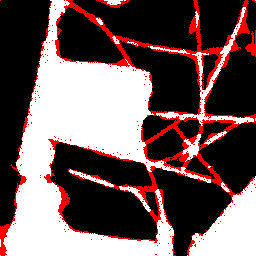}
        \caption{}
    \end{subfigure}%
    \begin{subfigure}[t]{.2\linewidth}
        \centering
        \includegraphics[width=.99\textwidth]{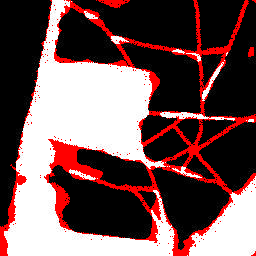}
        \caption{}
    \end{subfigure}%
    \begin{subfigure}[t]{.2\linewidth}
        \centering
        \includegraphics[width=.99\textwidth]{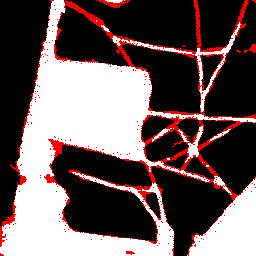}
        \caption{}
    \end{subfigure}%
    \begin{subfigure}[t]{.2\linewidth}
        \centering
        \includegraphics[width=.99\textwidth]{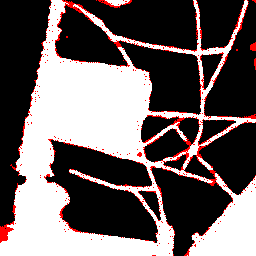}
        \caption{}
    \end{subfigure}%
    \begin{subfigure}[t]{.2\linewidth}
        \centering
        \includegraphics[width=.99\textwidth]{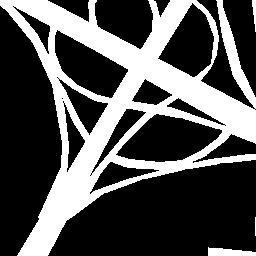}
        \caption{}
    \end{subfigure}%
    \vspace{1mm}
    \centering \text{Case-II}\\
    
    \captionof{figure}{Comparison of different state-of-the-art change detection methods on \textbf{CDD} dataset: \textbf{(a)} Pre-change image, \textbf{(b)} Post-change image, \textbf{(c)} FC-EF, \textbf{(d)} FC-Siam-diff, \textbf{(e)} FC-Siam-conc, \textbf{(f)} DT-SCN, \textbf{(g)} BIT, \textbf{(h)} ChangeFormer, \textbf{(i)} DDPM-CD (ours), and \textbf{(j)} Ground-truth. \textit{Note that true positives (change class) are indicated in white, true negatives (no-change class) are indicated in black, and false positives plus false negatives indicates in red.}}
    \label{fig:visual_cdd}
\end{figure}

\subsection{Qualitative change detection results}
Besides the quantitative results, we visually present predicted change maps to highlight the effectiveness of the proposed method compared to state-of-the-art methods. Figure \ref{fig:visual_levir}, Figure\ref{fig:visual_whu}, Figure \ref{fig:visual_dsifn}, and Figure \ref{fig:visual_cdd} display qualitative examples corresponding to the LEVIR-CD, WHU-CD, DSIFN-CD, and CDD datasets, respectively. In these visualizations, we represent the change class (positive class) in white, the no-change class (negative class) in black, and incorrectly predicted areas (false positives and false negatives) in red. Therefore, fewer red areas in a method indicate better performance in predicting both change and no-change classes.

For the LEVIR-CD dataset presented in Figure \ref{fig:visual_levir}. The first example depicts three building changes, while in the second case, many buildings have appeared, resulting in numerous building changes. In the first case, we can observe that our DDPM-CD accurately captures all three building changes, while other methods like FC-EF, FC-Siam-diff, FC-Siam-conc, BIT, and Changeformer either miss the building in the left-middle or can only partially predict the changes. When considering the second case, although most previously proposed change detection methods can predict most of the building changes, the predictions of DDPM-CD are more accurate and have fewer red areas.

For the WHU-CD dataset shown in Figure \ref{fig:visual_whu}, one with multiple building changes and the other with two building changes. In the first example, we can see that the change predictions from our DDPM-CD are more accurate and have sharper edges, while all the other methods struggle to predict the changes appearing at the bottom and struggling to differentiate building shadows with actual building parts. In the second case, which contains a very large building change on the left, challenging to recognize, all the other methods missed it, but our method was at least able to partially predict the change. Additionally, there is another change at the top, which was not predicted by any of the previous methods except BIT. However, our method has predicted most of the change area in that region and performed better than the prediction from BIT.

Differing from building change detection, let's now consider the visual quality of predictions on general change detection datasets like DSIFN-CD and CDD. We showcase prediction results for two examples from the DSIFN-CD dataset in Figure \ref{fig:visual_dsifn}. The first case includes changes due to highway construction, while the other contains changes related to new buildings. Given the nature of highways with numerous narrow and curved parts, all other methods miss most of these changes because it's challenging to predict due to the similarities in colors between highways and forests. However, our method can easily differentiate between highway and forest regions, resulting in highly accurate change predictions. In the second example, several challenging-to-recognize building changes appear, and our method accurately detects these regions better than all other methods, particularly in the changes visible on the left.

We also present two examples from the CDD dataset in Figure \ref{fig:visual_cdd}. The first example exhibits changes in buildings and roadways. However, the post-change image was captured during the snow season, making those changes challenging to recognize and predict. As observed, all other methods struggle to capture these changes, but our method accurately predicts them. In the second example, the narrow roadways and buildings visible in the pre-change image disappear in the second image. While the building changes are clearly visible, the narrow roadways, obscured by the forest, are challenging to predict. While state-of-the-art methods predict building change areas, they face difficulties with the narrow and obscured roadways. However, our method accurately predicts these narrow roadways, resulting in a high-quality change map.

All of these qualitative comparisons underscore the effectiveness of our proposed DDPM-CD method compared to the existing state-of-the-art methods. Moreover, it demonstrates the extraordinary ability of DDPM to deliver robust and discriminative features that are useful in downstream applications like change detection.

\renewcommand{\tabcolsep}{7pt}
\begin{table*}[htb!]
    \centering
    \caption{\textbf{The ablation study on the timestep $t$ used to extract multi-timestep feature representations}. We show that combining feature representations belonging to multiple timesteps improves the change detection performance on the $\texttt{val}$-set of LEVIR-CD, WHU-CD, DSIFN-CD, and CDD.}
    \begin{tabular}{p{2cm}ccccccccccccccc}
		\toprule
        
		\multirow{2}{*}{\parbox[c]{\linewidth}{\centering Time step $t$}} &\multicolumn{3}{c}{\bf LEVIR-CD~\cite{LEVIR}} && 
		\multicolumn{3}{c}{\bf WHU-CD~\cite{WHU-CD}} &&
		\multicolumn{3}{c}{\bf DSIFN-CD~\cite{DSIFN}} &&
		\multicolumn{3}{c}{\bf CDD~\cite{CDD}}\\ 
		\cmidrule{2-4} \cmidrule{6-8} \cmidrule{10-12} \cmidrule{14-16}
		&{\bf F1}&{\bf IoU}&{\bf OA}&&
		{\bf F1}&{\bf IoU}&{\bf OA}&&
		{\bf F1}&{\bf IoU}&{\bf OA}&&
		{\bf F1}&{\bf IoU}&{\bf OA}\\
		\midrule
		5               &   89.71 & 81.35 & 99.15&&
		                    91.57 & 84.46 & 99.19&&
		                    93.87 & 88.39 & 96.09&&
		                    91.24 & 83.89 & 91.24\\
		50              &   90.66 & 82.90 & 99.23&&
		                    92.74 & 86.47 & 99.31&&
		                    94.17 & 88.99 & 96.29&&
		                    93.78 & 88.28 & 98.60\\
		100             &   90.50 & 82.65 & 99.21&&
		                    92.78 & 86.54 & 99.31&&
		                    94.95 & 90.39 & 96.77&&
		                    94.32 & 89.25 & 98.72\\
		150             &   90.08 & 81.95 & 99.18&&
		                    92.34 & 85.77 & 99.27&&
		                    94.59 & 89.74 & 96.54&&
		                    94.34 & 89.29 & 98.75\\
		50, 100         &   \bf 91.02 & \bf 83.52 & \bf 99.26&&
		                    \textcolor{blue}{\bf93.09} & \textcolor{blue}{\bf87.07} & \textcolor{blue}{\bf99.34} &&
		                    \textcolor{blue}{\bf94.51} & \textcolor{blue}{\bf89.61} & \textcolor{blue}{\bf96.51} &&
		                    \textcolor{blue}{\bf94.91} & \textcolor{blue}{\bf90.31} & \textcolor{blue}{\bf98.85}\\
		50, 100, 400   &    \textcolor{red}{\bf 91.26} & \textcolor{red}{\bf 83.92} & \textcolor{red}{\bf 99.28}&&
		                    \textcolor{red}{\bf 93.50} & \textcolor{red}{\bf 87.80} & \textcolor{red}{\bf 99.38}&&
		                    \textcolor{red}{\bf95.38}& \textcolor{red}{\bf91.18} & \textcolor{red}{\bf94.05} &&
		                    \textcolor{red}{\bf95.64} & \textcolor{red}{\bf91.64} & \textcolor{red}{\bf99.00}\\
		50, 100, 650   &    \textcolor{blue}{\bf91.10} & \textcolor{blue}{\bf 83.67} & \textcolor{blue}{\bf 99.26}&&
		                    \bf 93.02 & \bf 86.95 & \bf 99.33&&
		                    \bf 95.07 & \bf 90.62 & \bf 96.87&&
		                    \bf 95.24 & \bf 90.90 & \bf 98.92\\
		\bottomrule
	\end{tabular}
    \label{tab:ablation}
\end{table*}
\section{Ablation study on multi-timestep features}
\label{sec:ablation}
This ablation study investigates the impact of utilizing different multi-timestep features ($t \in [0, T]$) from the diffusion model on change detection performance. We fine-tune the change detection classifiers using features obtained at various timesteps $t$ from the diffusion model to identify the timestep range that provides optimal semantics for change detection. Table \ref{tab:ablation} illustrates how change detection performance on the validation set varies when utilizing features sampled from different timesteps: $t=5, 50, 100, (50 \text{ and } 100), (50, 100 \text{ and } 400)$, and $(50, 100, \text{ and } 650)$ as inputs for training the hierarchical change classifier.

Our observations indicate that the most favorable change detection performance across all datasets is achieved when utilizing feature representations sampled within the range of $t \in [100, 400]$. Moreover, combining feature representations from multiple time samples, such as $t=50, 100$, and $400$, further enhances change detection performance. Consequently, we designate feature representations sampled at $t=$50, 100, and 400 as the default configuration for multi-timestep features, which is employed to report results on the test sets of all datasets presented in Table \ref{tab:quant_res}.

\section{Comparison of Computational Complexity}
\renewcommand{\tabcolsep}{7pt}
\begin{table*}[tbh!]
    \centering
    \caption{Comparison of computational complexity of different methods. We consider pre-change and post-change images of size $256 \times 256$.}
    \begin{tabular}{lrrr}
        \toprule
        Method & Trainable Params. (M) & GLOPs & Inference Time (ms)\\
        \midrule
        SimSiam~\cite{chen2020simple} & 12.49 & 4.76 & 1.04 \\
        MoCo-v2~\cite{chen2020improved} & 11.24 & 4.76 & 1.92 \\
        DenseCL~\cite{wang2021dense}&11.69 & 4.76 & 2.66\\
        CMC~\cite{cmc}&22.48 & 4.66 & 1.55\\
        SeCo~\cite{manas2021seasonal}&12.16&9.52&3.62\\
        \hline
        \rowcolor{gray}
        \hspace{3pt} DDPM & 390.95 & 716.40 & 28.75 \\
        \rowcolor{gray}
        \hspace{3pt} CD w/ $n=1$ & 39.08 & 25.99 & 1.85\\
        \rowcolor{gray}
        \hspace{3pt} CD w/ $n$=2 & 43.96 & 30.56 & 2.46\\
        \rowcolor{gray}
        \hspace{3pt} CD w/ $n$=3 & 46.41 & 32.84  & 2.56\\
        \rowcolor{gray}
        DDPM-CD (n=1) & $39.08$ & $1\times716.49+25.99=742.48$ & $1\times28.75+1.85=30.6$\\
        \rowcolor{gray}
        DDPM-CD (n=2) & $43.96$ & $2\times716.49+30.56=1458.97$ & $2\times28.75+2.46=59.35$\\
        \rowcolor{gray}
        DDPM-CD (n=3) & $46.41$ & $3\times716.49+32.84=2175.46$ & $3\times28.75+2.56=88.10$\\
        \hline
        \bottomrule
    \end{tabular}
    \label{tab:compute_complexity}
\end{table*}
Table \ref{tab:compute_complexity} compares the computational complexity of the proposed DDPM-CD with the existing methods. We benchmark our method for pre-change and post-change images of spatial resolution $256\times256$ and use an NVIDIA Quadro RTX 8000 GPU.

Our DDPM has a total of 390.95 million trainable parameters. The hierarchical change classifier has 39.08 million parameters if single-timestep features are used, 43.96 million parameters if two timesteps are used, and 46.41 million if three time-steps are used. Since we fine-tune only the hierarchical change classifier and keep the DDPM frozen, the total trainable parameters during the fine-tuning come from the hierarchical change detector. DDPMs usually require a higher number of parameters to enable their modeling capability, and more recent DDPMs have even higher parameter counts.

When considering GLOPs and inference time, the DDPM consumes 716.40 GLOPs per image pair and takes about 28.75 ms per image pair for one step forward pass. Since we utilize DDPM for feature extraction during fine-tuning and inference, it requires 1-3 forward passes to extract multi-step features, whereas if we use it in the synthesis, which usually involves 1000s of time-steps, it requires $\times 1000$ times. For our best model, which utilizes features corresponding to three time steps, it requires $3 \times 716.40$ = 2149.2 GLOPs and takes $28.75 \times 3$ = 86.25 ms. The hierarchical change detector, which processes those features and outputs a change map, requires 32.84 GLOPs and takes 2.56 ms when utilize features of three timesteps. Therefore, for the best model, it requires a total of 2149.2 + 32.84 = 2182.04 GLOPs and takes 86.25 + 2.56 = 88.81 ms.

In comparison to other state-of-the-art methods, our method exhibits higher counts of trainable parameters, GLOPs, and inference time. This observation is understandable because the DDPM necessitates a large network to enable its modeling power, allowing it to accurately capture the training distribution, unlike other architectures. We believe that despite the higher number of parameters and GLOPs, the final performance of our method outweighs these metrics when compared to other state-of-the-art methods. Exploring ways to reduce its model size while retaining its modeling capabilities and decreasing inference time would be both intriguing and timely. Presently, the current trend in diffusion models leans toward larger sizes, a direction driven by the demanding nature of handling extremely complex input data distributions, the need for high-quality image synthesis, and the increasing complexity of multi-modal data in the natural image domain.

\section{Conclusion}
\label{sec:conclusion}
In this paper, we proposed a novel remote sensing change detection method based on denoising diffusion probabilistic models. The proposed method first pre-trains an unconditional denoising diffusion probabilistic model on millions of off-the-shelf remote sensing images available to the public to learn the key semantics in the remote sensing images. The pre-trained diffusion model is then utilized to extract multi-scale feature representations from a given satellite image to train a lightweight change detection classifier. The proposed method differs from the existing deep change detection methods mainly in two aspects: (1) The pre-training of diffusion probabilistic model, which serves as the feature extractor in our case, does not require any labeled data.  This is valuable especially for remote sensing applications because most of the publicly available millions of remote sensing data are unlabeled or challenging to annotate. (2) Unlike existing deep change detection algorithms, we keep the weights of the trained diffusion model  frozen and only optimize the parameters of the lightweight change detection classifier, which helps efficient and fast deployment on the new change detection applications or datasets. Above all these advantages of the proposed approach, it also achieves significantly better performance compared to the SOTA change detection methods making it the new SOTA method for remote sensing CD.

\ifCLASSOPTIONcaptionsoff
  \clearpage
\fi
\bibliographystyle{IEEEtran}
\bibliography{egbib.bib}
\end{document}